\documentclass[preprint]{elsarticle}

\usepackage{graphicx}

\usepackage{amssymb}

\usepackage{lineno}

\usepackage{comment}
\usepackage{amsmath}
\usepackage{lscape}
\usepackage{multicol}
\usepackage{array}
\usepackage{txfonts}
\usepackage{multirow}
\usepackage{url}
\usepackage{float}
\newcolumntype{P}[1]{>{\centering\arraybackslash}p{#1}}
\newcolumntype{M}[1]{>{\centering\arraybackslash}m{#1}}

\usepackage[pdftex]{color}
\usepackage{xcolor}
\usepackage{colortbl}
\definecolor{Gray}{gray}{0.7}
\definecolor{Blue}{HTML}{0000FF}
\definecolor{Orange}{HTML}{FFD4A1}
\usepackage[font=footnotesize,labelfont=bf]{caption}
\usepackage[font=footnotesize,labelfont=bf]{subcaption}

\journal{}

\begin{document}

\begin{frontmatter}


\title{Blockchain-based Crowdsourced Deep Reinforcement Learning as a Service}



\author[add1]{Ahmed Alagha}
\author [add2,add3]{Hadi Otrok}
\author [add2,add3]{Shakti Singh}
\author[add2,add3]{Rabeb Mizouni}
\author[add2,add4,add1]{Jamal Bentahar}

\address[add1]{Concordia Institute for Information Systems Engineering, Concordia University, Montreal, QC, Canada}
\address[add2]{Department of Computer Science, Khalifa University, Abu Dhabi, UAE}
\address [add3]{Center of Cyber Physical Systems (C2PS), Khalifa University, Abu Dhabi, UAE}
\address [add4]{6G Research Center, Khalifa University, Abu Dhabi, UAE}
 
\begin{abstract}
Deep Reinforcement Learning (DRL) has emerged as a powerful paradigm for solving complex problems. However, its full potential remains inaccessible to a broader audience due to its complexity, which requires expertise in training and designing DRL solutions, high computational capabilities, and sometimes access to pre-trained models. This necessitates the need for hassle-free services that increase the availability of DRL solutions to a variety of users. To enhance the accessibility to DRL services, this paper proposes a novel blockchain-based crowdsourced DRL as a Service (DRLaaS) framework. The framework provides DRL-related services to users, covering two types of tasks: DRL training and model sharing. Through crowdsourcing, users could benefit from the expertise and computational capabilities of workers to train DRL solutions. Model sharing could help users gain access to pre-trained models, shared by workers in return for incentives, which can help train new DRL solutions using methods in knowledge transfer. The DRLaaS framework is built on top of a Consortium Blockchain to enable traceable and autonomous execution. Smart Contracts are designed to manage worker and model allocation, which are stored using the InterPlanetary File System (IPFS) to ensure tamper-proof data distribution. The framework is tested on several DRL applications, proving its efficacy.
\end{abstract}

\end{frontmatter}


\section{Introduction}
\label{Sec: Intro}
%
%
%
%

The recent advancements in Deep Reinforcement Learning (DRL) have demonstrated significant potential in applications such as robot swarms \cite{SHURRAB2023100867}, autonomous driving \cite{antonio2022multi}, and video games \cite{berner2019dota}. DRL is a paradigm within the realm of Artificial Intelligence (AI) that seamlessly blends the power of Deep Learning (DL) and Deep Neural Networks (DNNs) with the principles of Reinforcement Learning (RL). In DRL, agents are able to develop their own intelligence through rewarded interactions with the environment, based on a previously designed learning algorithm that translates the collected experiences into an action policy. 

Despite its significant success, designing and training DRL solutions has been, and still is, a challenging task. From one side, designing DRL solutions for real-world problems requires tremendous expertise and domain knowledge. Designers often have to go through intricate steps which include the modeling of the environment and its interactions, the modeling of the agents' observations, the choice of the policy optimization algorithm and the tuning of its hyperparameters, and the design of the reward function, to name a few \cite{li2017deep}. Each of these steps could significantly affect one another as well as the outcome of the learning process, and require customized solutions tailored to each application. These obstacles are further amplified with the typical challenges faced in DRL, which include sample inefficiency, curse of dimensionality, reward sparsity, and the exploration-exploitation dilemma \cite{gronauer2022multi}. On the other side, training DRL requires a significant amount of computational resources that might not be available for certain users or businesses. This is mainly due to the complexity of DNNs and DRL algorithms, and the high dimensionality of typical DRL environments. For example, the authors in \cite{berner2019dota} used nearly 51 thousand CPUs and 512 GPUs to train a DRL system for the game of Dota, and such resources are not accessible to the majority of users and researchers.

Over the past few years, the paradigm of Machine Learning as a Service (MLaaS) came as a way to promote greater accessibility to Machine Learning (ML) solutions for developers, business, and the general public. MLaaS is an offering that provides ML capabilities and infrastructure as a service to users in exchange for money \cite{wang2023b}. Generally, MLaaS providers target a wide spectrum of users, ranging from those with no experience in ML, to those who have the expertise but lack computational resources. In the first case, users interact with ML APIs by providing training data, which is used by the service to produce a trained model \cite{pan2021pnas}. The automated process usually involves data pre-processing, model training, and evaluation. Such services often cover a range of ML applications, including but not limited to classification, regression, clustering, and natural language processing. Service providers include Google AutoML and Vertex AI, which enable developers with limited ML expertise to train high-quality models. In the second case, MLaaS provides infrastructure and tools for experienced users to build and develop models, as well as computational resources, if needed. Such services include Amazon SageMaker, Microsoft Azure, and Google Colab.

Extending MLaaS directly into the realm of DRL is infeasible with the existing services. Firstly, DRL commonly requires much longer time and more computational resources to train when compared to typical ML problems. This would induce significant overhead on existing MLaaS providers, which usually operate via a shared platform between users that is allocated on demand \cite{ribeiro2015mlaas}. Secondly, the environment variability in DRL results in a highly dynamic, and often unpredictable, interaction between the agent and the environment during the training process. This contrasts with most ML scenarios in which the environment is static or well-defined. Thirdly, having few MLaaS providers entails higher costs on users, especially those with problems requiring high computational resources and long training times. Finally, and most importantly, the design and training processes in DRL are much more complex when compared to ML, hence requiring more human expertise and interventions. Since ML problems generally rely on static and available datasets, the learning process can be automated. This is unlike DRL, which relies on dynamic and sequential interactions with the environment that need to be modeled and optimized by an expert. Recent works in the literature propose crowdsourcing systems for ML tasks, with the aim of using the scalability, diversity, and expertise of the crowd \cite{aly2022pay, dong2022improving, washington2020precision, zhang2024instance, martin2023strong}. Crowdsourcing refers to the practice of obtaining services (to requesters) by soliciting contributions from a group of people (workers) \cite{elsokkary2023crowdsourced,alagha2020rfls, liang2021novel}. However, all the existing works in ML only crowdsource tasks related to data collection, data annotation, and model validation. To our knowledge, none of the existing methods discuss the crowdsourcing of ML or DRL training and model sharing tasks, which require expertise and computational capabilities in addition to DRL-related attributes. In summary, the drawbacks in existing methods are:
\begin{enumerate}
    \item Existing MLaaS systems are usually centralized and automated services, making them infeasible to DRL tasks that have higher computational overhead and require more distributed solutions and long training times.
    \item Having few MLaaS platforms entails higher costs and low accessibility to solutions.
    \item The lack of diverse expertise in existing MLaaS solutions hinders them incapable of addressing the complex and diverse DRL tasks that require intricate modeling and optimization by experts.
    \item Existing crowdsourcing systems for ML focus on data gathering and model validation tasks, with no consideration for training and model sharing tasks that require computational capabilities and expertise.
\end{enumerate}

To circumvent the aforementioned issues, this paper introduces a novel blockchain-based crowdsourced DRL as a service (DRLaaS) framework. In this work, the proposed framework addresses the lack of diverse expertise and computational capabilities in MLaaS by crowdsourcing DRL training tasks instead of relying solely on centralized platforms. It also introduces DRL-related worker recruitment for crowdsourcing, which existing ML crowdsourcing systems do not consider. The proposed framework covers two types of tasks, namely DRL training and model sharing. In DRL training tasks, users have the ability to request the full design and training of DRL solutions based on the specified requirements to be executed by the recruited workers. Such a task depends heavily on the computational capabilities and expertise of the worker. In model sharing tasks, users can benefit from pre-trained and available models by expert workers, which can be used to assist in training new DRL solutions using methods in knowledge transfer, such as Imitation Learning-based DRL and Demonstration Cloning \cite{alagha2022target, nair2018overcoming}. Here, the reputation of a worker and the suitability of the models they provide are instrumental when allocating pre-trained models to requesters. For both types of tasks, the proposed framework is responsible for managing the allocation of training tasks and models to suitable workers. For DRL training tasks, specific DRL-related metrics are designed to assess candidate workers according to the task requirements and based on their capabilities. Specifically, attributes like expertise in designing and training DRL solutions, reputation in accepting and finishing DRL tasks, and computational capabilities are considered in a Quality of Service (QoS) metric that is optimized using a greedy algorithm. The computational capability parameters are designed to assess critical worker attributes like GPU availability, RAM capacity, and CPU parallelization, which are pivotal for effective DRL training. For model sharing tasks, in addition to the worker's expertise and reputation in managing this type of tasks, a DRL model similarity metric is designed to assess the available models, which considers the DRL environment of the model and how suitable it is based on the requester's requirements. The framework is deployed on a Consortium Blockchain, which manages the interaction between requesters and workers, the task and resource allocation, and the model sharing processes. Blockchain is used, instead of centralized cloud-servers, to mitigate issues related to single-point of failure and the need of a single trusted server, by providing a decentralized, transparent, and autonomous platform with no repudiation. A Consortium Blockchain, specifically, provides better privacy, scalability, and efficiency when compared to public blockchains \cite{kadadha2022context}, and better allowance for collaboration and data sharing between entities when compared to private Blockchains. Simple and efficient Smart Contracts (SCs) are designed to manage the task and model allocation processes, and the InterPlanetary File System (IPFS) is used to manage the storage of models. In summary, the contributions of this work are as follows:

\begin{enumerate}
    \item The design of a comprehensive framework for crowdsourced DRLaaS that utilizes the expertise and computational capabilities of expert workers in answering tasks related to the design, training, and sharing of DRL solutions, aiming to assist inexperienced and experienced users in return for incentives.
    \item The design of task and model allocation processes using greedy methods that consider DRL-related and workers-related attributes, such as computational capabilities, task requirements, DRL environment details, model quality, and workers expertise and reputation.
    \item The design of DRL-related computational capability and model similarity attributes. Computational capability attributes consider the effect of RAM, CPU, and GPU specifically on the DRL training task. Model similarity attributes consider features in the DRL environment to assess how existing models can be beneficial to the requesting user.
    \item The design of simple and efficient smart contracts that fully manage the aforementioned processes on the blockchain, with the assistance of IPFS to store trained models. 
\end{enumerate}

\begin{figure}[H]
    \centering
    \includegraphics[width=0.8\columnwidth]{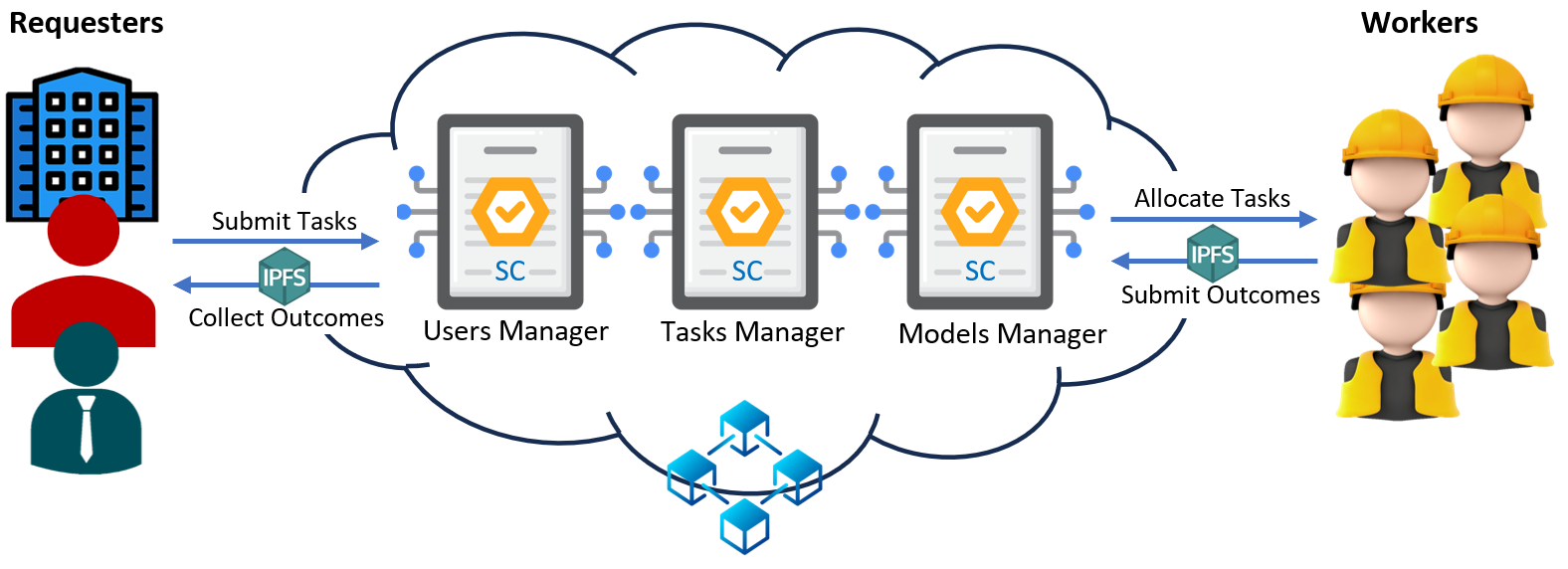}
    \caption{A general overview of the proposed framework.}
    \label{fig:GeneralOverview}
\end{figure}

A general overview of the proposed DRLaaS framework is shown in Fig. \ref{fig:GeneralOverview}. Requesters submit tasks by interacting with the smart contracts on the blockchain. The smart contracts manage users' registrations, tasks allocations, and model sharing processes. Once tasks are allocated and executed by workers, the returned outcomes (trained models) are shared via IPFS. The tasks' outcomes are then shared back to the requesters, and workers are paid accordingly.

The proposed framework is evaluated for several complex DRL applications, including Target Localization \cite{alagha2022target}, Autonomous Vehicles Fleet Coordination \cite{xidias2016path}, and Multi-Agent Maze Cleaning \cite{jiang2021multi}. 

The remainder of this paper is organized as follows: Section \ref{section: Related Work} reviews recent work in MLaaS and crowdsourcing for ML. Section \ref{section: DRL Formulation} gives background about DRL and highlights the important aspects of designing and training DRL solutions. Section \ref{Framework} formulates the problem and details the proposed blockchain-based DRLaaS framework. Experiments and results are presented and discussed in Section \ref{section: Sim & Eval}, and the paper is concluded in Section \ref{Conclusion}.

\section{Related Work}
\label{section: Related Work}

Since the proposed DRLaaS framework is the first to tackle the DRL problem in a crowdsourcing manner, this section gives an overview of existing works in the fields of 1) MLaaS and 2) Crowdsourcing for Machine Learning, which lay the foundations of the proposed framework.

\subsection{Machine Learning as a Service}
The advent of MLaaS provided several platforms in the industry that allow users with varying skill sets to leverage its capabilities. Several MLaaS providers offer pre-built ML models and/or tools that enable MLaaS users to obtain ML models. Google Cloud is one platform providing users with services, such as AutoML and Vertex AI. AutoML provides a user-friendly interface for users with limited experience, where users can upload their labeled datasets and the service takes care of model training and optimization. AutoML offers services for a varying set of tasks, including image recognition, natural language processing, and video analysis. Vertex AI, on the other hand, is an end-to-end ML platform that offers a set of tools for building and training ML models. Similar to AutoML, Amazon SageMaker and Microsoft Azure facilitates the end-to-end ML process by offering a range of tools that cater to different skill levels. Google Colab provides a cloud-based development environment for ML through a Jupyter Notebook interface. Colab is mainly known for providing access to powerful, but limited without paid subscription, GPUs and TPUs (Tensor Processing Units) for faster training. While SageMaker, Azure, and Colab can be used to train DRL systems, the higher complexities of DRL problems hinder their usability, especially for inexperienced users.

In terms of research, MLaaS attracted many scientists over the past few years. The proposal in \cite{ribeiro2015mlaas} is amongst the first works addressing MLaaS, where the authors present an architectural design for a MLaaS platform. In the proposed design, the data is received and processed by a Data Gatherer, and then a model is built and trained by a Modeler Composite. The authors in \cite{zhao2018packaging} propose Acumos, a platform capable of packaging ML models into portable containerized microservices, which can be easily integrated into different business applications. The main aim is to reduce the technical burden on developers when applying ML models to their applications. The authors in \cite{kumar2022sclera} propose a privacy-preserving MLaaS on resource-constrained devices at the
pervasive edge. The proposed framework uses methods such DNN splitting and quantization, enclave parallelization, and resource-aware offloading policies to protect clients' private data while using computing resources in the pervasive edge ecosystem. The authors in \cite{singh2022machine} propose a MLaaS framework that optimizes the allocation of limited ML resources by intelligently considering attributes such as service profile, region-wise resource usage patterns, and current ML resource usage. In \cite{graur2022cachew}, the authors propose Cashew, a service for ML data processing that caches common input data pipelines across jobs from different clients to optimize training throughput.

The aforementioned platforms and research proposals prove efficient automating and increasing accessibility to ML. However, they cannot be adapted to DRL problems due to the higher complexity of such problems and the lack of human expertise to design and train DRL solutions. 

\subsection{Crowdsourcing for Machine Learning}
The intersection between Crowdsourcing and ML is twofold: some works utilize ML and DRL in designing crowdsourcing systems \cite{abououf2020artificial, ren2022privacy, abououf2021machine}, while other works crowdsource certain ML tasks \cite{aly2022pay, puttinaovarat2020flood, washington2020precision, chang2017revolt, martin2023strong}. We present an overview of the second set of works in the literature, since our work aims to crowdsource DRL tasks. Existing works using crowdsourcing for ML mainly use the crowd for data collection and processing. In \cite{aly2022pay}, a dataset of breathing and coughing sounds that are collected via crowdsourcing is used for COVID-19 diagnosis. The authors in \cite{puttinaovarat2020flood} propose a ML framework for flood forecasting systems using data collected through crowdsourcing, where workers report data about actual flooding incidents, such as rainfall intensity levels, continuing rainfall duration, and drainage ability. The authors in \cite{washington2020precision} develop a ML model to evaluate the performance of workers in categorizing neurotypical and autistic children. The data are collected through a crowdsourcing platform, where workers watch videos of children and fill out a series of questions about the child's behavior. In \cite{chang2017revolt}, the authors propose a collaborative crowdsourcing system for ML data labeling. In the proposed system, groups of workers collaborate in labeling data through three stages: Vote (choosing the label), Explain (reasoning behind the label), and Categorize (review other workers' explanations). The authors in \cite{martin2023strong} propose a crowdsourced annotation framework for data in sound event detection applications. The goal is to estimate strong labels, i.e. data labels with high confidence, using weak labels that go through methods such as majority voting.

Despite the numerous research put into crowdsourcing data-related tasks for ML, there are no proposals, to our knowledge, that crowdsource the ML (or DRL) training process. This work aims to propose a comprehensive framework that addresses this issue by utilizing the expertise of the crowd in training DRL solutions.

\section{Background}
\label{section: DRL Formulation}
This section gives a general formulation of the DRL problem, along with the main aspects to be considered when designing and training DRL solutions.

\subsection{DRL Formulation}
In RL, agents interact with the environment based on their observations, and learn to alter their behaviors in response to rewards received \cite{sutton2018reinforcement}. In this setup, an autonomous agent observes a state $\boldsymbol{s_t}$ from its environment at time step $t$, and interacts with the environment by taking action $\boldsymbol{a_t}$. Based on this action, the environment returns a reward value $\boldsymbol{r_t}$, and transitions into a new state $\boldsymbol{s_{t+1}}$. The goal for the agent is to learn a policy $\boldsymbol{\pi}$ that maximizes the collected rewards by trying to take the best sequence of actions. A policy takes as input the agent's observations and returns the best action or a probability distribution over the possible actions. The recent advancements in DNN have led to their integration in DRL, where most DRL policies come in the form of DNNs. Formally, DRL is described as a Markov Decision Process (MDP), which consists of:
\begin{itemize}
    \item $\boldsymbol{\mathcal{S}}$ denotes the finite set of states;
    \item $\boldsymbol{\mathcal{A}}$ denotes the finite set of actions;
    \item $\boldsymbol{\mathcal{P}}$ is a set of Markovian state transition probabilities, where $\boldsymbol{\mathcal{P}(s_{t+1}|s_t,a_t)}$ is the probability that taking an action $\boldsymbol{a_t}$ in state $\boldsymbol{s_t}$ results in a transition to state $\boldsymbol{s_{t+1}}$;
    \item $\boldsymbol{\mathcal{R}(s_t, a_t, s_{t+1})}$ denotes a reward function;
    \item and $\boldsymbol{\gamma \in [0, 1]}$ is the discount factor for the trade-off between instantaneous and future rewards.
\end{itemize}
Generally, the policy $\boldsymbol{\pi}$ maps a state to a probability distribution over the possible actions $\boldsymbol{\pi : \mathcal{S} \rightarrow P(\mathcal{A} = a | \mathcal{S})}$. In most real-world problems, an agent does not have a full observation of the environment. In such cases, the problem is defined as a Partially Observable MDP (POMDP), where a policy maps the agent's observations to actions \cite{gronauer2022multi}.

\subsection{DRL Design and Training Requirements}
\label{subsection: DRL design}
The design and training of DRL models are complex tasks, requiring specialized \textit{expertise}, \textit{computational capabilities}, and often \textit{compatible pre-trained models}. This section explains these needs, which are to be met by the proposed crowdsourcing framework.

\subsubsection{Expertise: Environment, Reward, and Optimization}

The process of designing and training DRL solutions requires expertise in several areas. This process can be summarized in three main steps, namely 1) Environment Design, 2) Reward Engineering, and 3) Policy Optimization. The \textit{Environment Design} step is concerned with modeling the problem of interest in an environment that follows the DRL formulation. This requires the designer to have expertise in developing environments that mimic real-world dynamics and encapsulate the possible interactions (actions) between the agent(s) and the environment. For example, in a simple environment for the game of chess, the designer needs to fully understand the rules that govern the game, which are needed to define the possible actions for the chess agent. Additionally, the state of the chess environment could be modeled as a grid image of the board, or by numerical features representing the locations of the pieces. Such a choice is critical to be made by the designer, as it affects the training process. The \textit{Reward Engineering} step is critical since the behavior of the agents in DRL is directly influenced by the reward function. Reward functions require considerable engineering, which if done poorly could lead to local optima \cite{nair2018overcoming}. This requires the designer to have knowledge in the domain of interest, as well as expertise in reward shaping \cite{sami2023reward, sami2022graph}. For the game of chess, for example, a simple reward function is to assign a positive value only for a checkmate. However, this complicates the learning since the agent does not get regular feedback, and hence calls for the need for more complex reward functions that require expertise in this domain. In the \textit{Policy Optimization} step, the aim is to design policies that translate the agent's observations into actions, which are then optimized using the collected rewards. In DRL, policies are represented as DNNs, and the choice of architecture depends on several attributes, such as the type and dimensionality of the input and the problem complexity, which require expertise in DL. On the other hand, the optimization algorithm determines how the collected experiences (observations and rewards) update and optimize the agent's policy. Designing and choosing the appropriate optimization algorithm requires expertise in the domain of the application, as well as in hyperparameter tuning, which is essential in finding optimal solutions. For example, in the aforementioned chess environment, convolutional neural networks (CNNs) could be used as policy architectures if the state is represented as an image, while feed forward networks (FFNs) could be used if the state is represented in numerical features. These could be optimized using methods such as Proximal Policy Optimization (PPO), Deep Q-Networks (DQN), or Deep Deterministic Policy Gradients (DDPG). All of the aforementioned are decisions to be made by the designer.

\subsubsection{Computational Capabilities}

Due to the complexity of DRL, the training process usually requires heavy computational resources. While the choice of optimization algorithm has a significant effect on the learning speed and convergence, the availability of computational resources holds equal significance. For example, the availability of GPUs could significantly speed up the execution of DNNs, which in turn speeds up the learning, especially for image-based applications (2D data with Convolutional Neural Networks). Additionally, many policy optimization algorithms for DRL can harness parallel processing on multiple processing units (CPU cores or GPUs) to speed up the learning process by parallelizing the experience collection process. Moreover, most DRL optimization methods come with many several hyperparameters, which should be fine-tuned for each application. This process requires heavy computational resources, in addition to expertise in this domain.

\subsubsection{Model Availability and Compatibility}

The DRL training process has been recently approached using methods from imitation learning. Recent works \cite{alagha2023multi, damani2021primal, nair2018overcoming} proposed using demonstrations from an expert in guiding the learning of DRL agents. An expert model is a pre-trained model that is already familiar with the environment (or with a similar environment), and can help the agents collect better experiences, or can be used partially in Behavioral Cloning (BC) to enhance the DRL policy. Here, the availability of such expert models becomes an important issue to address. Additionally, the compatibility of the expert models with the current environment of interest is another crucial challenge. If the demonstrations suggested by the expert introduce variances to the current DRL training problem, this introduces difficulties in the learning convergence. This necessitates the availability of models that are similar to the current environment.

All the aforementioned needs are to be met by the proposed crowdsourcing framework in this work. The complicated design processes call for the need of expertise, which can be obtained through crowdsourcing. Additionally, crowdsourcing could help in providing computational resources through workers that have access to multiple CPUs and GPUs. The recent methods enhancing DRL through expert demonstrations call for the need of compatible expert models, which can also be obtained through crowdsourcing.

\section{Blockchain-based DRLaaS Framework}
\label{Framework}
\begin{figure}[h]
    \centering
    \includegraphics[width=0.53\columnwidth]{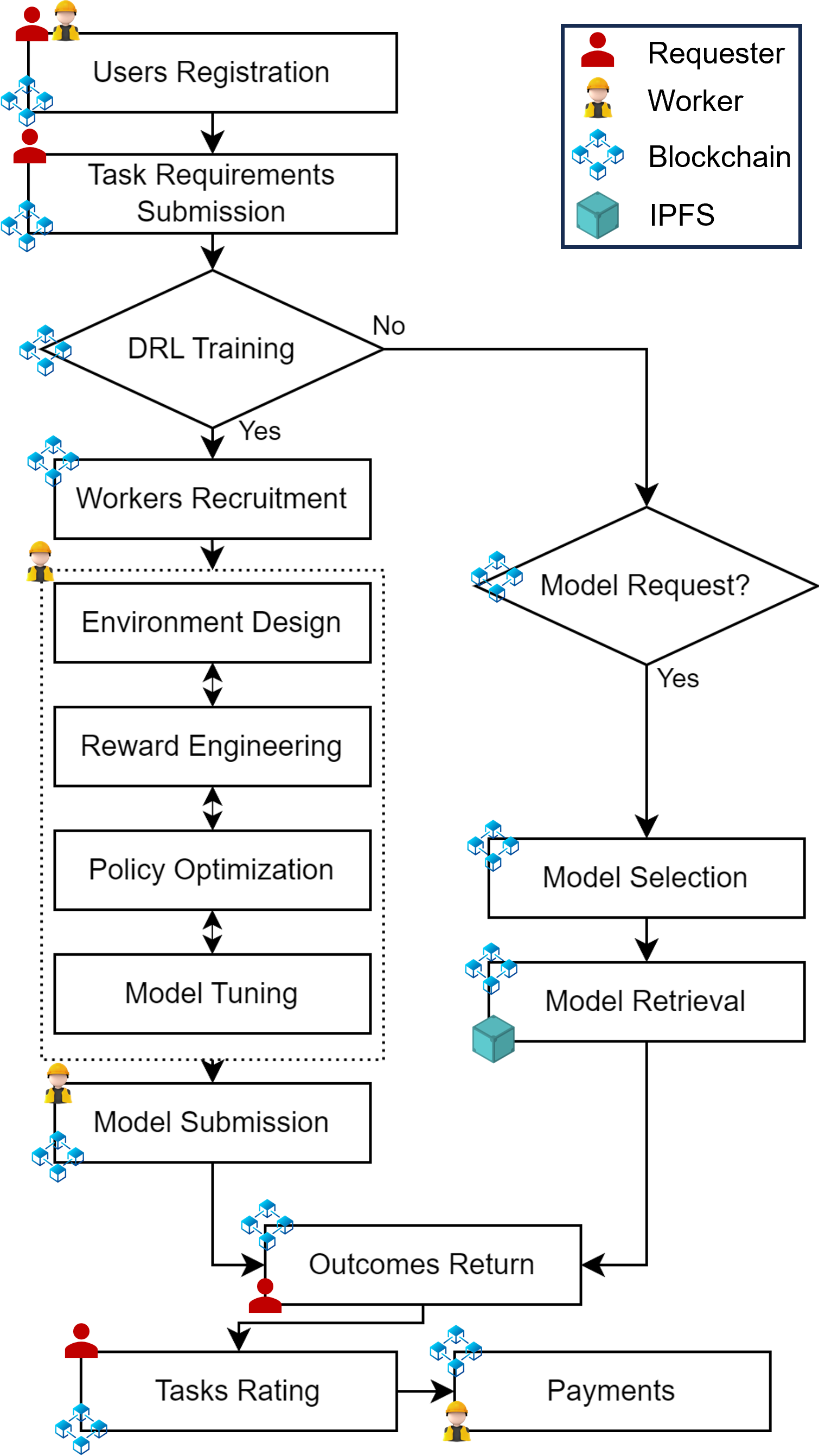}
    \caption{The proposed Blockchain-assisted DRLaaS framework. The different steps could involve a single entity, or an interaction between two entities. Entities include requesters, workers, the blockchain, and IPFS.}
    \label{fig:flowchartFramework}
\end{figure}

The proposed framework aims to target two types of DRLaaS tasks that can be requested by users, namely DRL training and model sharing. In DRL training tasks, users can request the design and training of DRL solutions to be done by expert workers, i.e. workers with experience in tackling problems in the domain of interest using DRL. This task is highly dependent on the worker's expertise and computational capabilities, as discussed in Section \ref{subsection: DRL design}. On the other hand, in model sharing tasks, users can request pre-trained models in certain domains, which can be shared by other expert workers. This task depends on the efficiency and compatibility of the available models with the requester's environment. The flow of the proposed framework is shown in Fig. \ref{fig:flowchartFramework}, which is to be detailed in this section by discussing the different DRLaaS tasks, the recruitment metrics and processes, and the design of the smart contracts. In this framework, users register in the system by submitting information about their attributes. Users can then submit task requests by indicating their requirements. For DRL training tasks, workers are selected following the designed recruitment process. The recruitment is done based on the workers' attributes and their expected Quality of Service (QoS), which considers metrics related to the expertise and computational capabilities of the candidate workers. A recruited worker then performs the training task, which includes steps such as environment design, reward engineering, policy optimization, and model tuning, before submitting the resultant model. For DRL model sharing, workers initially declare the availability of DRL models by submitting information about their models to the platform. Once a model request is submitted, a selection mechanism uses these information to compute the QoS for the candidate workers based on their attributes, as well as their model's attributes, to assess their efficiency and compatibility. When a worker is selected, they are tasked with sharing their model through the IPFS. In all the tasks, once the workers finish execution, the platform returns the outcomes to the requesters and forwards payments to workers. Subsequently, requesters rate the tasks by submitting reviews, which are used in the future to compute the QoS of a worker, as will be discussed in this section.

\subsection{Problem Formulation}

Each task type, i.e. DRL training and model sharing, has its own requirements and recruitment metrics to be used when selecting workers. Generally, given a set of tasks $T=\{T_1, T_2, T_3, ...\}$ submitted by requesters and a pool of candidate workers $W=\{W_1, W_2, W_3, ...\}$, the aim is to select a set of workers $W^{T_i} = \{W^{T_i}_1, W^{T_i}_2, W^{T_i}_3,... \}$ for each task $T_i$. Each task $T_i$ is defined as a tuple in the form of $T_i=(ID^{T_i}, NW^{T_i}, K^{T_i}, D^{T_i}$ $, Rep^{T_i}_{min}, R^{T_i}_{min}, Dom^{T_i}, TC^{T_i}, CR^{T_i})$, where $ID^{T_i}$ is the task ID, $NW^{T_i}$ is the number of workers desired for task $T_i$ , $K^{T_i}$ is the task type (training or sharing), $D^{T_i}$ is a detailed description of the problem of interest, $Rep^{T_i}_{min}$ and $R^{T_i}_{min}$ are the minimum reputation and rating required for a worker to perform the task, $Dom^{T_i}$ is the task domain, $TC^{T_i}$ is the time constraint specified for the task to be completed within, and $CR^{T_i}$ is the computational requirements for the task. On the other hand, a worker $W_j$ is defined as a tuple in the form $W_j = (ID^{W_j}, Rep_{Tr}^{W_j}, Rep_{MS}^{W_j}. R_{Tr}^{W_j}, R_{MS}^{W_j}, Dom^{W_j}, Exp^{W_j}_{Tr}, Exp^{W_j}_{MS}, CC^{W_j})$, where $ID^{W_j}$ is the worker's ID, $Rep_{Tr-Dom^{T_i}}^{W_j}$ and $Rep_{MS-Dom^{T_i}}^{W_j}$ are the worker's reputations based on historical performances in the platform for the two task types in domain $Dom^{T_j}$, $R_{Tr-Dom^{T_i}}^{W_j}$ and $R_{MS-Dom^{T_i}}^{W_j}$ are the worker's ratings for the two task types in domain $Dom^{T_j}$, $Dom^{W_j}$ is the set of domains that the worker is familiar with, $Exp_{Tr-Dom^{T_i}}^{W_j}$ and $Exp_{MS-Dom^{T_i}}^{W_j}$ are the worker's expertise in the two task types in domain $Dom^{T_j}$, and $CC^{W_j}$ is the worker's computational capabilities. It is worth mentioning that a worker's reputation is mainly based on their commitment to accepting and finishing tasks before the specified deadlines. On the other hand, a worker's rating is based on their performance, obtained through feedback on the trained models from users that have already used the service. Table \ref{Table: attributes} summarizes the task and worker attributes and their definitions.

The aforementioned attributes are utilized differently based on the task type. For each of the tasks, the following section describes the task requirements and worker recruitment process.

\subsection{Worker Recruitment Parameters}

Each of the two task types requires a dedicated worker recruitment process to select the most suitable set of workers to execute the task. To address this, we propose two different Quality of Service (QoS) metrics which are used to assess candidate workers and select the most suitable for each type of tasks. The proposed QoS metrics take into consideration the task requirements and worker attributes, and ensure that the selected workers meet the desired expectations.

\subsubsection{DRL Training Tasks}
\label{subsection DRL Training}
A worker here is asked to design a suitable environment for the task and engineer an efficient reward function. Additionally, the worker is required to choose a suitable DRL optimization algorithm, train a policy for the task at hand, and fine-tune the trained model (policy). The worker's expertise is essential for designing the environment and choosing a suitable policy architecture (type of neural network) and optimization method based on the provided environment. The availability of computational resources is also crucial for training and fine-tuning the model in a reasonable time.

\renewcommand{\arraystretch}{1.5}
\begin{table}[H]
\caption{List of attributes and their definitions.}
\vspace{-1.3em}
\setlength{\tabcolsep}{3pt}
\begin{center}
\begin{tabular}{|P{50pt}|p{220pt}|}
\hline
\rowcolor{Gray}
Attribute & Definition\\
\hline
$W^{T_i}$ & Set of workers for task $T_i$\\ 
\hline
$ID^{T_i}$ & ID of task $T_i$\\ 
\hline
$NW^{T_i}$ & Number of workers desired for task $T_i$\\ 
\hline
$K^{T_i}$	& Type of task $T_i$\\ 
\hline
$D^{T_i}$ & Description of task $T_i$\\ 
\hline
$Rep^{T_i}_{min}$ & Minimum reputation requirement for task $T_i$\\ 
\hline
$R^{T_i}_{min}$ & Minimum rating requirement for task $T_i$\\ 
\hline
$Dom^{T_i}$ & Domain of task $T_i$ (DRL training or model sharing)\\ 
\hline
$TC^{T_i}$	& Time constraint of task $T_i$\\ 
\hline
$CR^{T_i}$ & Computational requirements for task $T_i$\\ 
\hline
$ID^{W_j}$	& ID of worker $W_j$\\ 
\hline
$Dom^{W_j}$ & Set of domains covered by  worker $W_j$\\ 
\hline
$Rep^{W_j}_{Tr-Dom^{T_i}}$	& Reputation of worker $W_j$ in DRL training for $Dom^{T_i}$\\ 
\hline
$Rep^{W_j}_{MS-Dom^{T_i}}$	& Reputation of worker $W_j$ in model sharing for $Dom^{T_i}$\\ 
\hline
$R^{W_j}_{Tr-Dom^{T_i}}$	& Rating of worker $W_j$ in  DRL training for $Dom^{T_i}$\\ 
\hline
$R^{W_j}_{MS-Dom^{T_i}}$	& Rating of worker $W_j$ in model sharing for $Dom^{T_i}$\\ 
\hline
$Exp^{W_j}_{Tr-Dom^{T_i}}$	& Expertise of worker $W_j$ in DRL training for $Dom^{T_i}$\\ 
\hline
$Exp^{W_j}_{MS-Dom^{T_i}}$	& Expertise of worker $W_j$ in model sharing for $Dom^{T_i}$\\ 
\hline
$CC^{W_j}$	& Computational capabilities of worker $W_j$\\
\hline
\end{tabular}

\end{center}
\label{Table: attributes}
\end{table}
\renewcommand{\arraystretch}{1}

To quantify the computational capabilities of a worker, we consider three important computational resources that are essential for DRL training: Central Processing Unit (CPU), Graphics Processing Unit (GPU), and Random Access Memory (RAM). The availability of multiple CPU cores and a powerful GPU is crucial for DRL training, as discussed in Section \ref{subsection: DRL design}. Since the series/brand of the GPU is what matters the most, it is assumed that the proposed crowdsourcing platform accepts a pre-defined list of GPU series, which are to be checked against the one available for the worker in the constraints. Finally, during the process of DRL training, huge amounts of data in the form of experiences (observations and rewards) need to be stored, which requires high RAM storage.

For a given DRL training task $T_i$, the requester specifies a set of requirements and constraints, given as:

\begin{itemize}
    \item Problem Description ($D^{T_i}$): this requirement describes the detailed nature of the problem for task $T_i$, which is needed by the worker to design the DRL environment. This includes descriptions of the type of data available to collect, the type of interactions between the agent and the environment, the type of interactions between the agents themselves (in multi-agent systems), etc. This attribute comes in the form of a textual description. For example, in the problem of target search, where a group of UAVs try to find a certain target, the problem description contains information about the number of sensing agents, the number of targets (single/multi), the dynamicity of the environment (with or without obstacles), the type of data that can be collected by the agents (images or sensor readings), etc.
    \item Problem Domain ($Dom^{T_i}$): this requirement describes the general domain of the desired task $T_i$, which is essential to focus on candidate workers with experience in similar tasks within the same domain. For example, in the problem of target search, the problem domain could be "robot swarms", which encompasses a range of problems that demand similar expertise.
    \item Reputation Requirement ($Rep_{min}^{T_i}$): this requirement describes the minimum reputation required by the requester, to be met by the candidate workers for task $T_i$. Reputation is a measurement of the historical commitment of the worker, which is mainly based on acceptance and completion of previous tasks.
    \item Time Constraint ($TC^{T_i}$): this requirement specifies the time window within which task $T_i$ is to be completed.
    \item Number of Workers ($NW^{T_i}$): in some problems, a requester might desire the task to be performed by multiple independent workers, to increase the likeliness of efficient results.
    \item Minimum number of CPU cores $CR^{T_i}_{CPU}$.
    \item Minimum RAM capacity $CR^{T_i}_{RAM}$.
\end{itemize}

It is worth mentioning that, in most cases, the number of GPU cores does not significantly affect the learning process, hence only the availability of a suitable GPU will be checked, regardless of the number of cores. To assess the expertise of a candidate worker for a task of DRL training, assume $N_{Tr-Dom^{T_i}}^{W_j}$ is the number of DRL training tasks completed by worker $W_j$ in domain $Dom^{T_i}$, then their expertise $Exp_{Tr-Dom^{T_i}}^{W_j}$ for DRL training in that domain is computed as:

\begin{equation}
    Exp_{Tr-Dom^{T_i}}^{W_j} = \frac{N_{Tr-Dom^{T_i}}^{W_j}}{N_{Tr-Dom^{T_i}}^{max}}
    \label{ExpFormula}
\end{equation}
where $N_{Tr-Dom^{T_i}}^{max}$ is the maximum number of training tasks completed by one of the candidate workers in the same domain, which is used for normalization. This equation measures the expertise of a candidate worker relative to the pool of available workers, normalized between 0 and 1, where 1 is given to the candidate worker with the greatest number of tasks completed. On the other hand, the worker's reputation $Rep^{W_j}_{Tr-Dom^{T_i}}$ for DRL training in domain $Dom^{T_i}$ is characterized by two attributes, namely commitment rate $CM^{W_j}_{Tr-Dom^{T_i}}$ and completion rate $CP^{W_j}_{Tr-Dom^{T_i}}$. Assume $A^{W_j}_{Tr-Dom^{T_i}}$ is the total number of tasks assigned to worker $W_j$, $Ac^{W_j}_{Tr-Dom^{T_i}}$ is the total number of tasks accepted by worker $W_j$, and $E^{W_j}_{Tr-Dom^{T_i}}$ is the total number of tasks completed by worker $W_j$, all for DRL training in domain $Dom^{T_i}$, then $CM^{W_j}_{Tr-Dom^{T_i}}$ and $CP^{W_j}_{Tr-Dom^{T_i}}$ are given as:

\begin{equation}
    CM^{W_j}_{Tr-Dom^{T_i}} = \frac{Ac^{W_j}_{Tr-Dom^{T_i}}}{A^{W_j}_{Tr-Dom^{T_i}}}
    \label{CMFormula}
\end{equation}
 \begin{equation}
    CP^{W_j}_{Tr-Dom^{T_i}} = \frac{E^{W_j}_{Tr-Dom^{T_i}}}{Ac^{W_j}_{Tr-Dom^{T_i}}}
    \label{CPFormula}
\end{equation}

In these equations, $CM^{W_j}_{Tr-Dom^{T_i}}$ reflects the confidence the platform has in worker $W_j$ to accept the assigned task, while $CP^{W_j}_{Tr-Dom^{T_i}}$ reflects the confidence in worker $W_j$ to complete the tasks they accepted to perform. The reputation $Rep^{W_j}_{Tr-Dom^{T_i}}$ of the worker is computed using the geometric mean, which is given as:

\begin{equation}
    Rep^{W_j}_{Tr-Dom^{T_i}} = \sqrt{CM^{W_j}_{Tr-Dom^{T_i}} \times CP^{W_j}_{Tr-Dom^{T_i}}}
    \label{RepFormula}
\end{equation}

Another metric to be considered in the QoS is the worker's performance rating $R_{Tr-Dom^{T_i}}^{W_j}$ for DRL training tasks in the desired domain, which is based on reviews given to the worker by previous requesters. Here, reviews could be in the form of a numerical rating in a given scale, which are averaged to give the worker a score between 0 and 1. Finally, to quantify the computational capabilities of a worker $CC^{W_j}$, a metric is proposed based on the number of CPU cores. Assume that the numbers of CPU cores available for worker $W_j$ are given as $N_{CPU}^{W_j}$, then the worker's computational capabilities $CC^{W_j}$ is quantified as:

\begin{equation}
    CC^{W_j} = \frac{2}{\pi} tan^{-1}{(w_1 N_{CPU}^{W_j})}
\end{equation}

The $tan^{-1}{(x)}$ function is used because its output increases rapidly initially with $x$ then slows down as $x$ gets higher. This is essential to capture the effect of CPU cores on DRL training. The parallelization process in the training process significantly speeds up the learning. However, after a certain number of parallel environments (depending on the problem), further parallelization does not bring more benefit. Generally, more complex DRL tasks could benefit from more parallelization. The weight $w_1$ controls the stretch (the rapid increase) of the function. High values of these weights help differentiating between workers with low number of CPU cores, but workers with high number of cores would nearly get the same score (which is sufficient for less complex DRL tasks). On the other hand, low values help differentiating between workers with high number of cores (which is desirable for more complex DRL tasks). The $tan^{-1}{(x)}$ function is multiplied by $\frac{2}{\pi}$ to normalize the output between 0 and 1.

Considering all the aforementioned metrics, the QoS of worker $W_j$ for the DRL training task in domain $Dom^{T_i}$ is then given as:

\begin{equation}
    QoS^{W_j}_{Tr-Dom^{T_i}} =  Exp^{W_j}_{Tr-Dom^{T_i}} \times Rep^{W_j}_{Tr-Dom^{T_i}} \times R_{Tr-Dom^{T_i}}^{W_j} \times CC^{W_j}
    \label{QoSTrainFormula}
\end{equation}

In this QoS function, all the attributes have values between 0 and 1. The format of Eq. \ref{QoSTrainFormula} is common in crowdsourcing literature \cite{alagha2021sdrs, kadadha2022chain}, as it provides normalized metrics that are used to assess and compare candidate workers. The aforementioned attributes are updated regularly following task assignments and completion by the workers. The proposed QoS is to be maximized while meeting the constraints for each candidate worker, given as:
\begin{itemize}
    \item $Rep^{W_j}_{Tr-Dom^{T_i}} \geq Rep^{T_i}_{min}$
    \item $R^{W_j}_{Tr-Dom^{T_i}} \geq R^{T_i}_{min}$
    \item $Dom^{T_i} \in Dom^{W_j}$
    \item $N_{CPU}^{W_j} \geq CR_{CPU}^{T_i}$
    \item $N_{RAM}^{W_j} \geq CR_{RAM}^{T_i}$
    \item The brand and series of $W_j$'s GPU is in the accepted list.
\end{itemize}

The QoS, along with the task constraints, are then used in an optimization method to select most suitable workers, which is described later in Section \ref{SubSubSectionOptimization}.

\subsubsection{DRL Model Sharing Tasks}
In model sharing tasks, a user shares information about their pre-trained model with the platform, with the hope of receiving incentives if the model is requested by other users. For a given model sharing task $T_i$, a requester specifies the Problem Description ($D^{T_i}$), Problem Domain ($Dom^{T_i}$), Reputation Requirement ($Rep_{min}^{T_i}$), Time Constraint ($TC^{T_i}$), and Number of Workers ($NW^{T_i}$). In $D^{T_i}$, the requester specifies information about the environment they hope to train, which are essential to find suitable shared models. The number of workers indicates the desired number of models to be shared with the requester, as some DRL methods require multiple models to assist in the learning \cite{alagha2023blockchain}. 

For model sharing tasks in the framework, the worker's expertise $Exp_{MS-Dom^{T_i}}^{W_j}$, reputation $Rep_{MS-Dom^{T_i}}^{W_j}$, and performance rating $R_{MS-Dom^{T_i}}^{W_j}$ are defined similar to the attributes for the DRL training tasks in Section \ref{subsection DRL Training} (as per Eqs. \ref{ExpFormula}-\ref{RepFormula}). Additionally, it is essential to assess the similarity between the environment within which the worker's model has been trained and the environment of the requester. It is assumed that DRL problems in the same domain $Dom^{T_i}$ have a set of $d$ pre-defined environment attributes $A = [F_1, F_2, ..., F_d]$ that characterize the environment. Let $m^{W_j}_k$ be the $k^{th}$ model shared by worker $W_j$ (assuming a worker can share multiple models). If $m^{W_j}_k$ is within the task domain $Dom^{T_i}$, then the similarity $S$ between model $m$ and the requirements of task $T_i$ is given as:

\begin{equation}
        S(m_k^{W_j},{T_i}) = \sum_{n=1}^d w_n \times |F_n^m - F_n^{T_i}| \; , \; \; \; \; \sum_{n=1}^d w_n = 1
    \label{eq: Similarity}
\end{equation}
where $F^m$ is the set of environment attributes associated with shared model, and $F^{T_i}$ is the set of environment attributes required by the task. The main intuition behind this metric is that models from similar environments could be of more benefit to the requester. The nature of the environment attributes depends on the task, but some examples include the number of agents, number of obstacles (in obstacle avoidance tasks), number of destinations (in autonomous vehicles applications), etc. Depending on the problem, some attributes might have more importance than others, and hence we use a weighted some with a weight $w_n$ for each attribute.

Considering all the aforementioned metrics, the QoS of worker $W_j$ for the DRL model sharing task is then given as:

\begin{equation}
\begin{split}
        QoS^{W_j}_{MS-Dom^{T_i}} = \frac{Exp^{W_j}_{MS-Dom^{T_i}} \times Rep^{W_j}_{MS-Dom^{T_i}} \times R_{MS-Dom^{T_i}}^{W_j}}{1 + S(m_k^{W_j},{T_i})}
\end{split}
\label{QoSModelSharing}
\end{equation}
where the +1 in the denominator is used for smoothing to avoid issues when $S(m_k^{W_j},{T_i})$ is zero, indicating that the two environments are exactly the same. The QoS is subject to the following constraints:
\begin{itemize}
    \item $Rep^{W_j}_{MS-Dom^{T_i}} \geq Rep^{T_i}_{min}$
    \item $R^{W_j}_{MS-Dom^{T_i}} \geq R^{T_i}_{min}$
    \item $Dom^{T_i} \in Dom^{W_j}$
\end{itemize}

\subsection{Recruitment Optimization Process}
\label{SubSubSectionOptimization}
\begin{figure}[H]
    \centering
    \includegraphics[width=0.6\columnwidth]{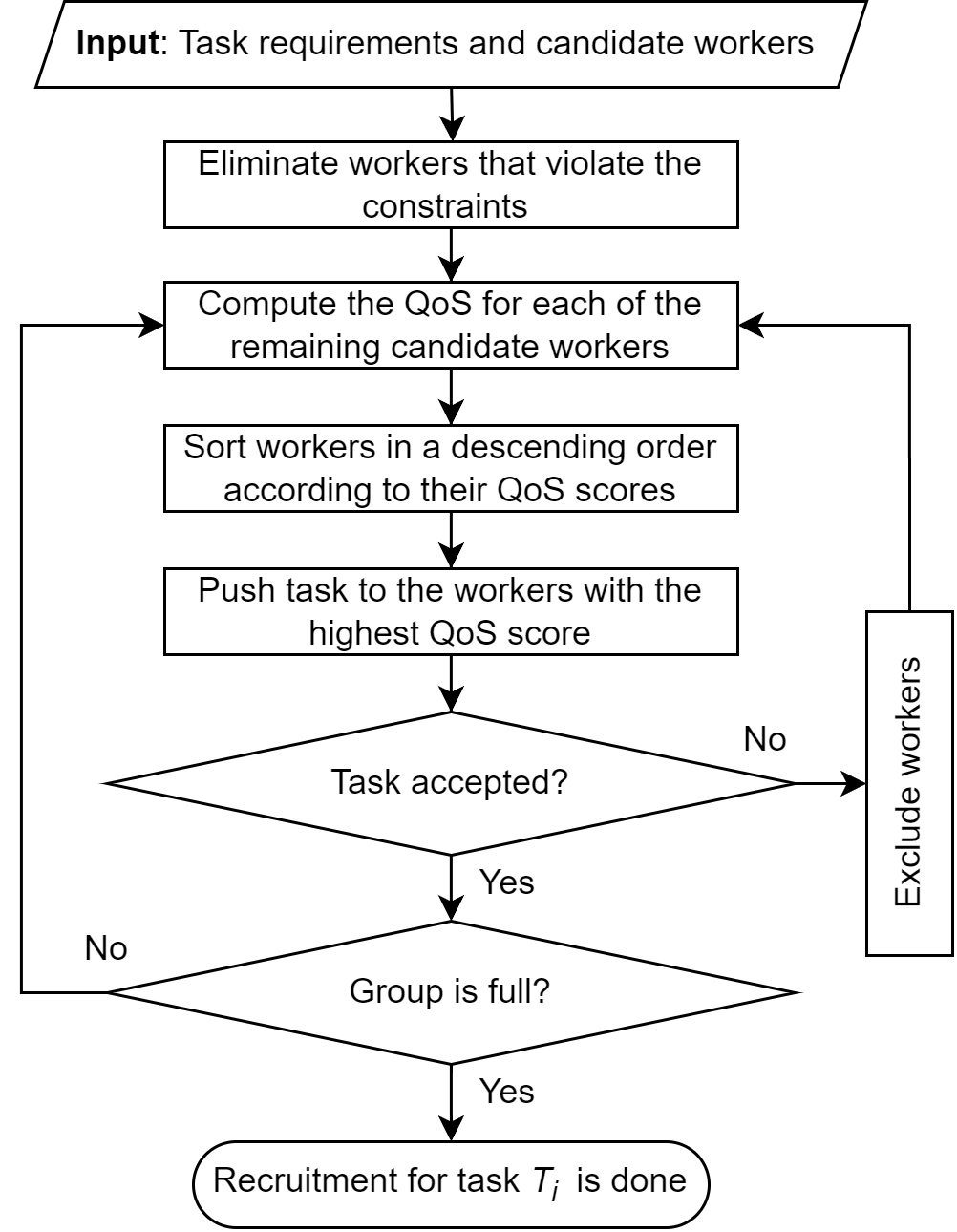}
    \caption{Flowchart of the recruitment optimization process.}
    \label{Fig:recruitment Flowchart}
\end{figure}

Given a task $T_i$ and a pool of candidate workers, the recruitment optimization process aims to recruit a group of $NW^{T_i}$ workers that maximize the expected QoS, while meeting the task requirements and constraints. To do so, a Greedy algorithm for worker recruitment is used, where the optimization process is treated as a knapsack problem. Greedy algorithms are the common in crowdsourcing recruitment works \cite{alagha2022influence, kadadha2022context}, due to their simplicity and scalability. While other algorithms, such as Genetic algorithm and Particle Swarm Optimization, could be used, the simplicity of Greedy algorithm makes it more suitable for deployment on the blockchain. In this problem, the aim is to maximize the weight of items filled in the knapsacks without violating its maximum capacity. In the context of worker recruitment, the maximum capacity represents the specified group size $NW^{T_i}$, while the weights represent the individual QoS values of the workers. Regardless of the task type (training or model sharing), the recruitment process is the same, with the QoS evaluation and task requirements being the difference between the task types. Once a task is pushed to a worker, they are given a time limit to accept, after which the task request is retracted and given to the next best worker. The recruitment process used in this work is described in Fig. \ref{Fig:recruitment Flowchart}.

\subsection{Smart Contracts Implementation}
In this work, the crowdsourcing platform is built on top of a Consortium Blockchain. The blockchain is responsible for managing users' registration, task requests, task allocation, and feedback through smart contracts. A blockchain is used instead of a centralized management system to provide a decentralized, transparent, and autonomous platform for crowdsourcing with no repudiation. A Consortium Blockchain, specifically, is used due to its ability to offer increased privacy, shared control, efficiency, cost savings, and trust for multiple organizations or entities collaborating on a project or sharing data \cite{kadadha2022chain}. A Consortium Blockchain is operated by a group of entities, which introduces increased privacy and trust when compared to public Blockchains, and more collaboration allowance when compared to private Blockchains, making them suitable for crowdsourcing. Recent works explored the utilization of blockchain with DRL \cite{alagha2023blockchain, sami2024learnchain}, where agents cooperate to train models on the chain, but none provided solutions or services for users. 

To execute the proposed crowdsourcing framework on the blockchain, three smart contracts are designed: 1) Users Manager Contract (UMC), Tasks Manager Contract (TMC), and Models Manager Contract (MMC). The users interact with UMC to register in the system by providing information. Task requesters interact with the TMC to submit tasks and provide requirements. The TMC is responsible for the worker recruitment, task allocation, task submission, and feedback processes. If users decide to share their trained models with the platform, they interact with the MMC, which manages and keeps track of the available DRL models. In all tasks, a worker could share the task outcomes, including trained models, through the InterPlanetary File System (IPFS). IPFS returns a unique Content Identifier (CID) that can be used to access the file. The IPFS is a protocol designed to create a content-addressable Peer-to-Peer (P2P) decentralized file system \cite{benet2014ipfs}. Workers share the CID with the smart contracts when returning the task outcomes.

The details of the \textbf{Users Manager Contract (UMC)} are shown in Table \ref{tab:UMC}. The \textit{Worker} data structure holds the worker's information. The information in this structure include the \textit{Worker Address} (Ethereum address) and their \textit{Reputation}. The \textit{Tasks Assigned} and \textit{Tasks Accepted} attributes reflect the total number of tasks assigned to and accepted by the worker. The \textit{Domains} attribute lists the domains covered by the worker, where each domain is represented by an index. The $Status$ indicates whether the worker is active or idle. The \textit{Expertise} indicates the number of tasks completed by the worker per each of the 2 task types. The \textit{Total Ratings} indicates the sum of ratings received for the tasks performed by the worker, for each of the 2 task types. The \textit{Comp. Capabilities} attribute represents the number of CPU cores, the RAM capacity, and the GPU type available for the worker. Finally, the \textit{Requester} data structure holds the \textit{Requester Address}.

\begin{table}[h]
    \caption{Users Manager Contract (UMC)}
    \vspace{-0.5em}
    \centering
    \begin{tabular}{|c|c|c|}
    \hline
    \rowcolor{Gray}
    \multicolumn{3}{|c|}{\textbf{Data Structure}}\\
    \hline
    \rowcolor{Orange}
    \multicolumn{3}{|c|}{\textbf{Worker}}\\
    \hline
    \textit{Worker Address }(address) & \multicolumn{2}{c|}{\textit{Reputation} (uint)}\\
    \hline
     \textit{Tasks Assigned} (uint) & \multicolumn{2}{c|}{\textit{Domains} (uint[])}\\
    \hline
    \textit{Tasks Accepted} (uint) & \multicolumn{2}{c|}
    {\textit{Status} (uint)}\\
    \hline
    \textit{Expertise} (uint[2]) & \multicolumn{2}{c|}{\textit{Total Ratings} (uint[2])}\\
    \hline
     \multicolumn{3}{|c|}{\textit{Comp. Capabilities} (uint[3])}\\
     \hline
    \rowcolor{Orange}
    \multicolumn{3}{|c|}{\textbf{Requester}}\\
    \hline
     \multicolumn{3}{|c|}{\textit{Requester Address }(address)}\\
    \hline
    \rowcolor{Orange}
    \multicolumn{3}{|c|}{\textbf{Variables}}\\
    \hline
    \multicolumn{3}{|c|}{\textit{Workers List} (address $\rightarrow$ Worker)}\\
    \hline
    \multicolumn{3}{|c|}{\textit{Domain Workers} (uint $\rightarrow$ address[])}\\
    \hline
    \rowcolor{Gray}
    \textbf{Function} & \textbf{Parameters} & \textbf{Return}\\
    \hline
    \textit{addWorker()} & Worker Information & -\\
    \hline
    \textit{addRequester()} & Requester Address & -\\
    \hline
    \textit{updateStatus()} & Status & -\\
    \hline
    \textit{updateInfo()} & Performance Details & -\\
    \hline
    \textit{getWorkers()} & Domain & \textit{Worker}[]\\
    \hline
    \end{tabular}
    \label{tab:UMC}
\end{table}

The UMC stores workers' information in the \textit{Workers List} mapping, which maps a worker's address to their \textit{Worker} object. Workers are grouped into domains in the \textit{Domain Workers} mapping, which maps a domain code to an array of Ethereum addresses for workers in that domain. The \textit{addWorker()} and \textit{addRequester()} functions allow users to register in the platform by providing necessary information to create \textit{Worker} or \textit{Requester} objects. The \textit{updateStatus()} and \textit{updateInfo()} functions are responsible for updating the worker's attributes frequently following events in the platform, like performing a task or receiving a review. The \textit{getWorkers()} function is responsible for retrieving all worker objects in given a domain code.

The \textbf{Task Manager Contract (TMC)} is shown in Table \ref{tab:TMC}. The \textit{Task} data structure holds information about the task, including the \textit{Requester}'s Ethereum address, the \textit{Duration} limit within which the task is to be finished, and the \textit{Type} of the task. The \textit{No. Workers} attribute reflects the number of workers required by the requester, \textit{Min. Reputation} and \textit{Min. Rating} describe the reputation and rating requirements for the task, \textit{Domain} indicates the domain code of the task, \textit{Problem Description} contains a textual description of the problem details, the \textit{Status} indicates whether the task is pending or completed, and \textit{Computational Reqs.} indicates the minimum requirements for CPU, GPU, and RAM. 

\begin{table}[H]
    \caption{Tasks Manager Contract (TMC)}
    \vspace{-0.5em}
    \centering
    \begin{tabular}{|c|c|c|}
    \hline
    \rowcolor{Gray}
    \multicolumn{3}{|c|}{\textbf{Data Structure}}\\
    \hline
    \rowcolor{Orange}
    \multicolumn{3}{|c|}{\textbf{Task}}\\
    \hline
    \textit{Requester} (address) & \multicolumn{2}{c|}{\textit{Duration} (uint)}\\
    \hline
     \textit{Type} (uint) & \multicolumn{2}{c|}{\textit{No. Workers} (uint)}\\
    \hline
    \textit{Min. Reputation} (uint) & \multicolumn{2}{c|}
    {\textit{Min. Rating} (uint)}\\
    \hline
    \textit{Problem Description} (string) & \multicolumn{2}{c|}{\textit{Status} (uint)}\\
    \hline
    \textit{Computational Reqs.} (uint[3]) & \multicolumn{2}{c|}{\textit{Domain} (uint)}\\
    \hline
    \rowcolor{Orange}
    \multicolumn{3}{|c|}{\textbf{Variables}}\\
    \hline
    \multicolumn{3}{|c|}{\textit{Domain Tasks} (uint $\rightarrow$ Task[])}\\
    \hline
    \rowcolor{Gray}
    \textbf{Function} & \textbf{Parameters} & \textbf{Return}\\
    \hline
    \textit{addTask()} & Task Information & -\\
    \hline
    \textit{allocateTask()} & Task Requirements & -\\
    \hline
    \textit{updateTaskStatus()} & Status & -\\
    \hline
    \textit{submitOutcome()} & Task Outcomes & -\\
    \hline
    \end{tabular}
    \label{tab:TMC}
\end{table}

The TMC stores the tasks information in a \textit{Domain Tasks} mapping, which maps a domain code to an array of task objects in that domain. The requester interacts with the \textit{addTask()} function by providing necessary information about the task to create a \textit{Task} object. The \textit{allocateTask()} function is responsible for worker recruitment and forwarding the task to the selected workers. the task status throughout the process is updated through the \textit{updateTaskStatus()} function, while the \textit{submitOutcome()} functions is called to submit the outcomes of a given task.  

The \textbf{Model Manager Contract (MMC)} is shown in Table \ref{tab:MMC}. The MMC is responsible for storing information about shared models. The \textit{Owner} attribute stores the address of the owner, the \textit{CID} stores the IPFS identifier for the shared files, \textit{Description} stores a textual description of the model and its application, \textit{Domain} indicates the domain code of the model's application, and \textit{Environment Details} stores attributes that identify the model's environment.

\begin{table}[H]
    \caption{Models Manager Contract (MMC)}
    \vspace{-0.5em}
    \centering
    \begin{tabular}{|c|c|c|}
    \hline
    \rowcolor{Gray}
    \multicolumn{3}{|c|}{\textbf{Data Structure}}\\
    \hline
    \rowcolor{Orange}
    \multicolumn{3}{|c|}{\textbf{Model}}\\
    \hline
    \textit{Owner} (address) & \multicolumn{2}{c|}{\textit{CID} (string)}\\
    \hline
     \textit{Description} (string) & \multicolumn{2}{c|}{\textit{Domain} (uint)}\\
    \hline
    \multicolumn{3}{|c|}
    {\textit{Environment Details} (uint[])}\\
    \hline
    \rowcolor{Orange}
    \multicolumn{3}{|c|}{\textbf{Variables}}\\
    \hline
    \multicolumn{3}{|c|}{\textit{Domain Models} (uint $\rightarrow$ Model[])}\\
    \hline
    \rowcolor{Gray}
    \textbf{Function} & \textbf{Parameters} & \textbf{Return}\\
    \hline
    \textit{addModel()} & Model Info & -\\
    \hline
    \textit{allocateModel()} & Model Requirements & Model[]\\
    \hline
    \end{tabular}
    \label{tab:MMC}
\end{table}

The MMC stores models' information in the \textit{Domain Models} mapping, which maps a domain code to the available models for that domain. A worker calls the \textit{addModel()} function to add details about the shared model, while the \textit{allocateModel()} function is responsible for finding the most suitable models amongst the available models based on the requester's requirements. 

In terms of time complexity, most of the functions in the three smart contracts have a complexity of O(N) or O(1). The \textit{addWorker()}, \textit{addRequester()}, \textit{addTask()}, and \textit{addModel()} functions have a complexity of O(N), where N is the number of existing elements (workers, requesters, etc.), since the functions check for duplicates before adding. Aside from the allocation functions, all the remaining functions have a simple complexity of O(1). The \textit{allocateTask()} and \textit{allocateModel()} functions employ a greedy algorithm that uses a sorting mechanism to find the best workers, which hence has a time complexity of O(N logN).

\subsection{Framework Time Sequence }

\begin{figure}[h]
    \centering
    \includegraphics[width=\textwidth]{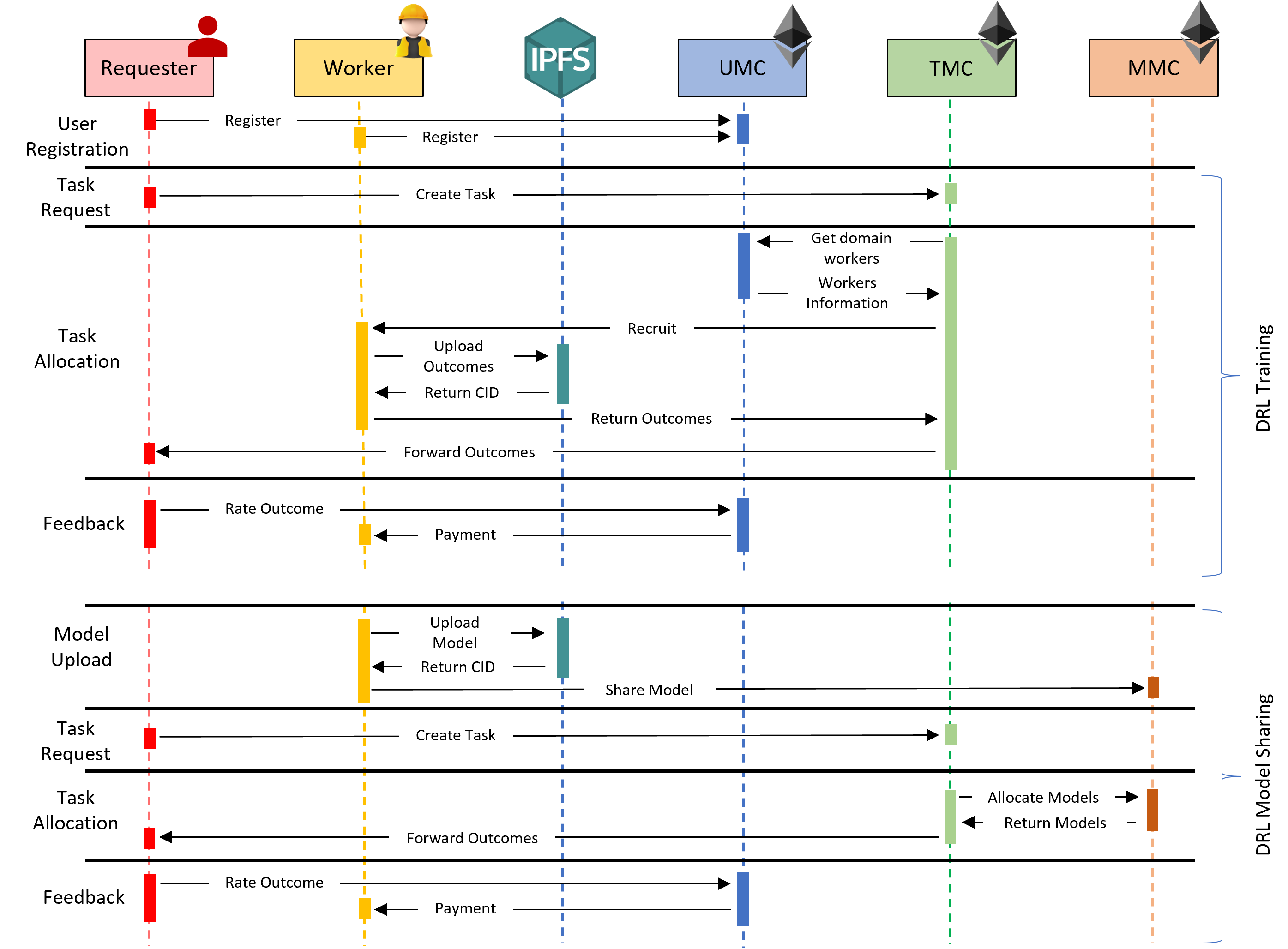}
    \caption{The interactions between the users and smart contracts as part of the proposed framework.}
    \label{fig:timesequence}
\end{figure}

Figure \ref{fig:timesequence} shows a time sequence diagram for scenarios under the proposed blockchain-based crowdsourced DRLaaS. It discusses the interactions between the users and the smart contracts constituting the framework. For DRL training tasks, these interactions are given as follows:

\begin{itemize}
    \item \textbf{User Registration}: Users register to the UMC by invoking the \textit{addWorker()} and \textit{addRequester()} functions. Each user, worker or requester, has a unique Ethereum address linked to their account. Workers provide information related to their domains and computational capabilities. The rest of the attributes are initialized and updated following task executions.
    \item \textbf{Task Request and Allocation}: A request creates a task by interacting with the TMC through the \textit{addTask()} function and providing the necessary information and the required payment. The TMC allocates the task to suitable workers through the \textit{allocateTask()} function, and workers perform the task and return their outcomes through IPFS to TMC through the \textit{submitOutcome()} function. The outcomes are then forwarded to the task requester.
    \item \textbf{Feedback}: Requesters rate the provided outcomes, and the ratings are used in the UMC to update workers' attributes. Workers then get paid for the tasks performed.
\end{itemize}

As for the DRL model sharing tasks, following the user registration, the process is as follows:
\begin{itemize}
    \item \textbf{Model Upload}: Workers who wish to share their trained models upload their files to the IPFS and interact with the MMC through the \textit{addModel()} function by sharing the model details.
    \item \textbf{Task Request and Allocation}: A requester creates a model sharing task by interacting with the TMC and specifying the task type and the attributes of the desired environment. The TMC interacts with the MMC by invoking the \textit{allocateModel()} function, which returns the most suitable models, which are then forwarded to the requester.
    \item \textbf{Feedback}: The requester rates the shared model, and the UMC updates the workers' attributes accordingly and provides the payments
\end{itemize}

\section{Simulation and Evaluation}
\label{section: Sim & Eval}
This section presents and discusses several experiments conducted to validate the proposed methods. The experiments are conducted on several Multi-agent DRL environments, namely Target Localization \cite{alagha2023multi,alagha2022target}, Fleet Coordination for Autonomous Vehicles \cite{xidias2016path}, and Multi-Agent Maze Cleaning \cite{jiang2021multi}. We opted to employ Multi-agent DRL instead of single-agent DRL due to its increased complexity, which allows for a more rigorous examination of the robustness and adaptability of our proposed methods. All the simulations have been conducted using an Intel E5-2650 v4 Broadwell workstation equipped with 128 GB RAM, 800 GB SSD, and NVIDIA P100 Pascal GPU (16 GB HBM2 memory).

\subsection{DRL Application Environments}
\begin{figure}[H]
    \vspace{-1em}
     \centering
     \begin{subfigure}{0.36\columnwidth}
         \centering
         \includegraphics[width=\linewidth]{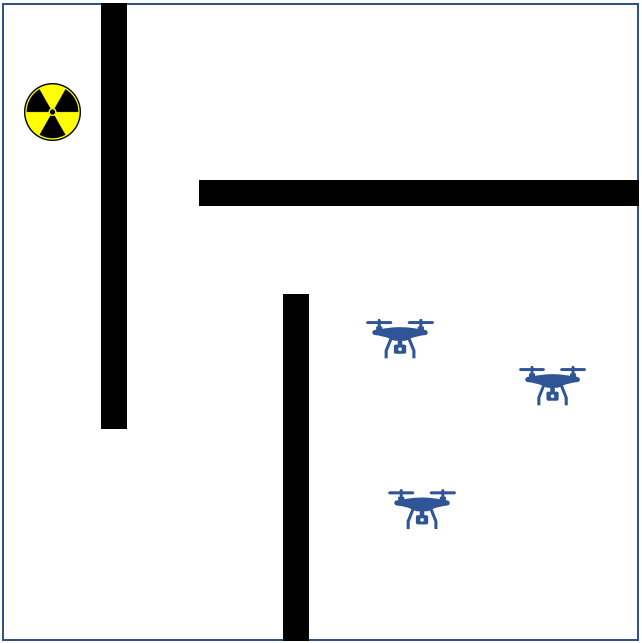}
         \caption{Target Localization}
         \vspace{1em}
         \label{Fig: TargetLocalization}
     \end{subfigure}
     \hspace{2em}
     \begin{subfigure}{0.3675\columnwidth}
         \centering
         \includegraphics[width=\linewidth]{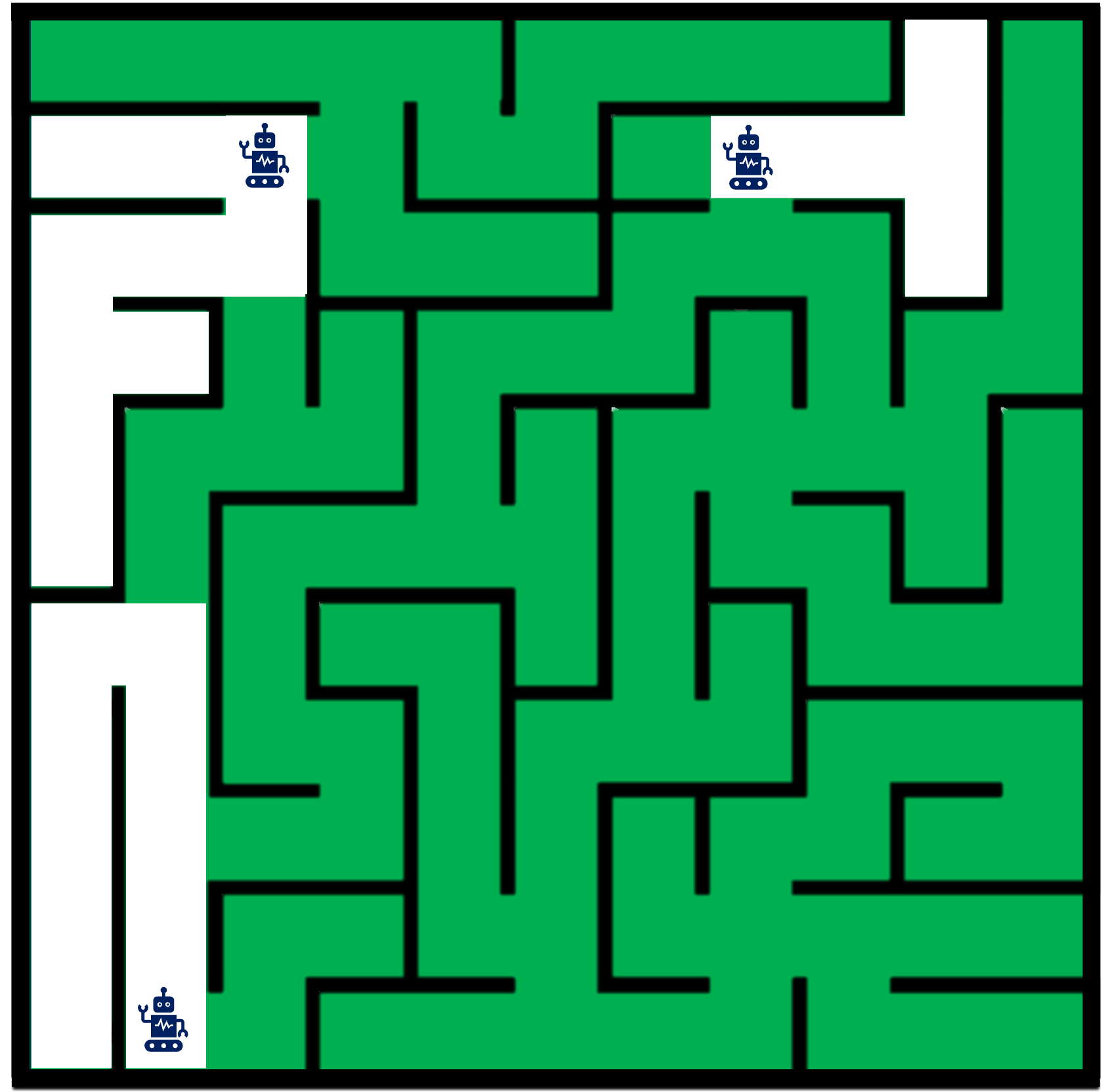}
         \caption{Maze Cleaning}
         \vspace{1em}
         \label{Fig: Maze Cleaning}
     \end{subfigure}
     \begin{subfigure}{0.40\columnwidth}
         \centering
         \includegraphics[width=\linewidth]{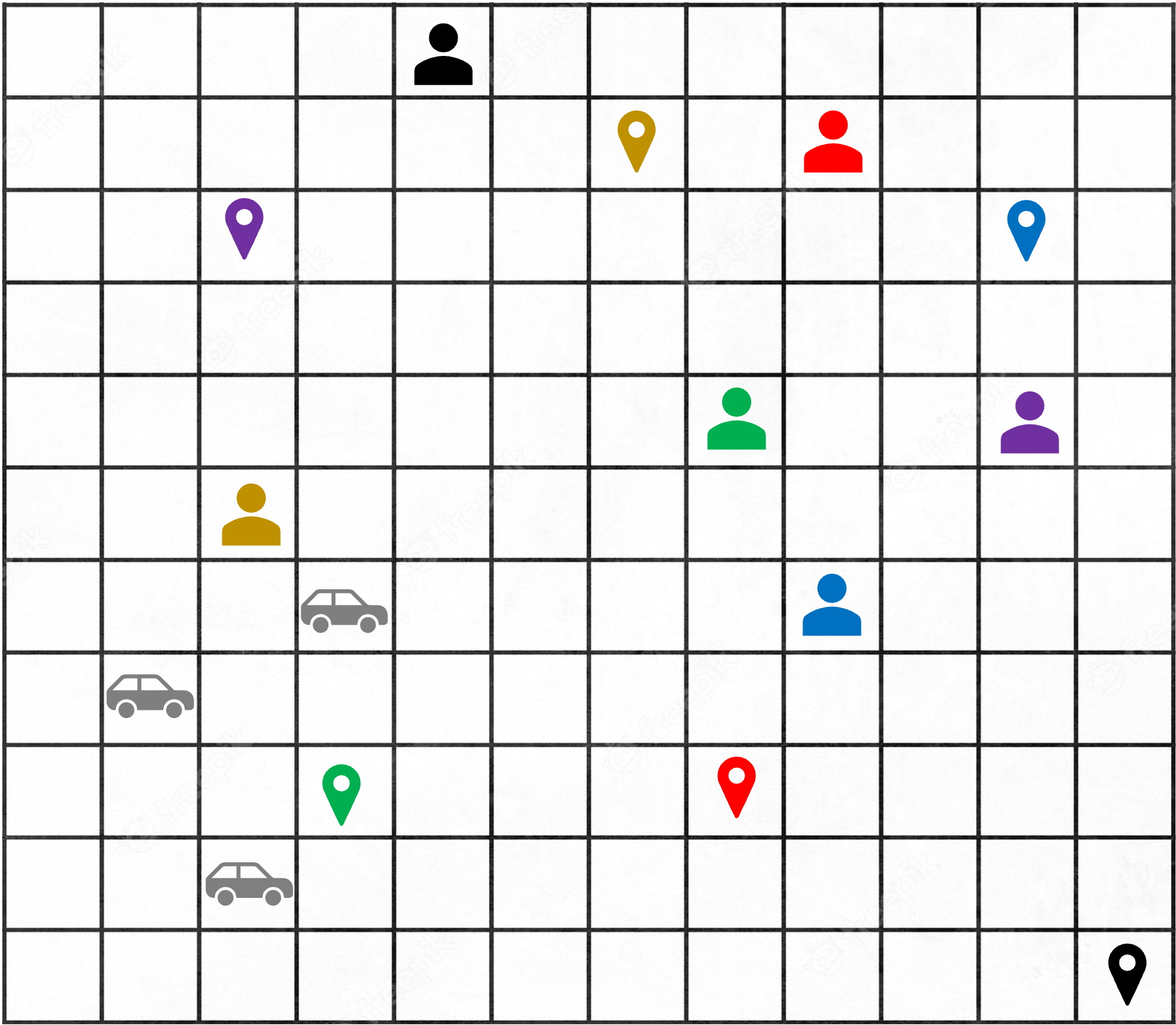}
         \caption{Fleet Coordination}
         \label{Fig: Fleet Coordination}
     \end{subfigure}
     \caption{Use-case scenarios of the DRL application environments used to validate the proposed methods.}
        \label{Fig: AppEnvironments}
\end{figure}

The DRL environments used in to validate the proposed methods are:
\begin{itemize}
    \item Target Localization \cite{alagha2023multi,alagha2022target}: this is a multi-agent problem in which the location of a certain target is to be identified, based on sensory readings collected by the sensing agents. This is common in applications related to radiation monitoring, search and rescue, and path-finding. In this problem, the sensing agents (robots or UAVs) observe the environment, collect data readings, and communicate with each other in order to cooperate and locate the unknown target. The learning problem is complicated since the agents need to learn how to communicate and coordinate, in addition to how to take the best set of actions in order to reach the target as fast as possible. We model the DRL environment as discussed and presented in \cite{alagha2022target}. The agents' observations are modeled as 2D heatmaps and fed to a Convolutional Neural Network (CNN) that acts as an actor network in a Proximal Policy Optimization (PPO) algorithm. A sample scenario of the target localization problem is shown in Fig. \ref{Fig: TargetLocalization}.
    \item Multi-Agent Maze Cleaning \cite{jiang2021multi}: in this problem, a group of agents is placed in a maze with the task of cleaning it as fast as possible. Initially, the maze is entirely dirty, and each spot covered by an agent is considered to have been cleaned. This problem requires coordination between the agents to allocate tasks, and for them to learn how to quickly clean the maze. Each agent observes its own location, the location of the other agents, as well as the status of the map, in 2D format. The observations are fed to a CNN as the actor network in a PPO algorithm. A sample scenario of the maze cleaning problem is shown in Fig. \ref{Fig: Maze Cleaning}.
    \item Fleet Coordination for Autonomous Vehicles \cite{xidias2016path}: in this problem, a team of autonomous vehicles is tasked with picking up and dropping customers at specific locations. Customers are randomly distributed in a certain area of interest, with each customer having a certain desired destination. Vehicles have a limit in terms of the number of customers accommodated simultaneously. The team's goal is to minimize the time needed to pick up and drop off customers off at their destinations. The complexity of the learning comes from the fact that the agents (i.e. the vehicles) need to cooperate to properly allocate tasks, such as the time is minimized. This is seen as a multi-objective DRL problem, since each agent is required to find the shortest path that goes through customers towards their destinations, while coordinating with other agents. The agents observe their locations, the customers' locations, as well as the desired destinations, and take actions accordingly. The observations are modeled as 2D heatmaps and fed to CNNs with PPO as the DRL optimization algorithm. A sample scenario of the fleet coordination problem is shown in Fig. \ref{Fig: Fleet Coordination}.
\end{itemize}

These environments are examples of complex DRL problems that require expertise and computational resources. In the following sections, the proposed methods will be validated using these environments, in which users rely on the proposed DRLaaS framework to push tasks to workers and get outcomes.

\subsection{Performance Analysis}
This section evaluates the methods in the proposed DRLaaS framework. To validate the proposed framework, the key attributes in the worker selection metrics are evaluated, including agent's computational capabilities $CC^{W_j}$ and model similarity $S(m_{k}^{W_j}, T_i)$. The learning convergence in the following experiments is evaluated in terms of Episode Length, which is the time it takes for the agents to finish the task. The episode length is tracked throughout the learning, to reflect how fast the agents learn to efficiently perform the task. It is also essential to track how long the training process takes, in wall time, in order to verify the importance of the proposed selection metrics. For all the experiments, the conditions of the simulations are fixed, and only the variables under examination are varied.

\subsubsection{DRL Training Tasks}

\begin{figure}[H]
     \centering
     \begin{subfigure}{0.49\columnwidth}
         \centering
         \includegraphics[width=\linewidth]{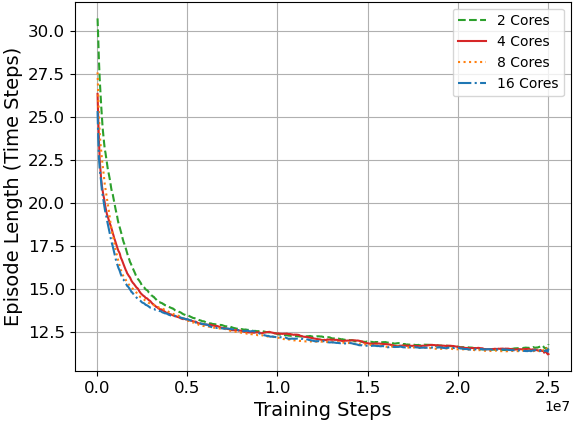}
         \caption{Target Localization}
         \label{Fig: TL_CPU}
     \end{subfigure}
     \begin{subfigure}{0.487\columnwidth}
         \centering
         \includegraphics[width=\linewidth]{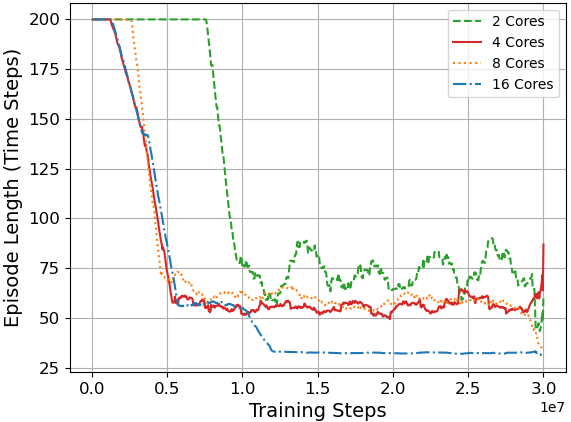}
         \caption{Maze Cleaning}
         \label{Fig: MC_CPU}
     \end{subfigure}
     \begin{subfigure}{0.49\columnwidth}
         \centering
         \includegraphics[width=\linewidth]{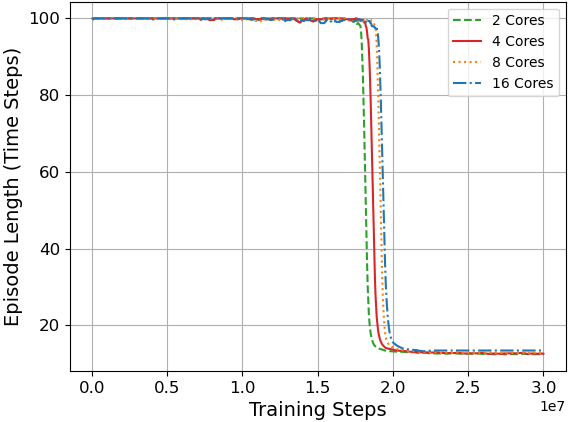}
         \caption{Fleet Coordination}
         \label{Fig: AV_CPU}
     \end{subfigure}
    \caption{The effect of parallelizing the DRL process over a varying number of CPU cores on the learning convergence, for different DRL environments.}
        \label{Fig: CPU_Results}
\end{figure}

As discussed in Section \ref{subsection DRL Training}, most common DRL algorithms utilize parallelized processing to run copies of the environment for experience collection during training. To verify the importance of considering the worker's CPU capabilities $CC_{CPU}^{W_j}$ during the recruitment stage for DRL training tasks, Fig. \ref{Fig: CPU_Results} presents the effect of varying the number of CPU cores on the training convergence of DRL for the 3 applications. For many applications, parallelizing the data collection in DRL does not significantly affect the learning convergence, which is the case for Target Localization (Fig. \ref{Fig: TL_CPU}) and Fleet Coordination (Fig. \ref{Fig: AV_CPU}). However, in some applications where the task has a long horizon (takes generally longer time steps to finish), increasing the number of parallel processes brings benefits, as seen in Fig. \ref{Fig: MC_CPU} for the Maze Cleaning environment. This is because parallelizing the data collection enhances the exploration process, since more unique experiences in different parallel environments are being collected, resulting in better training. 

\begin{figure}[H]
     \centering
      \begin{subfigure}{0.49\columnwidth}
         \centering
         \includegraphics[width=\linewidth]{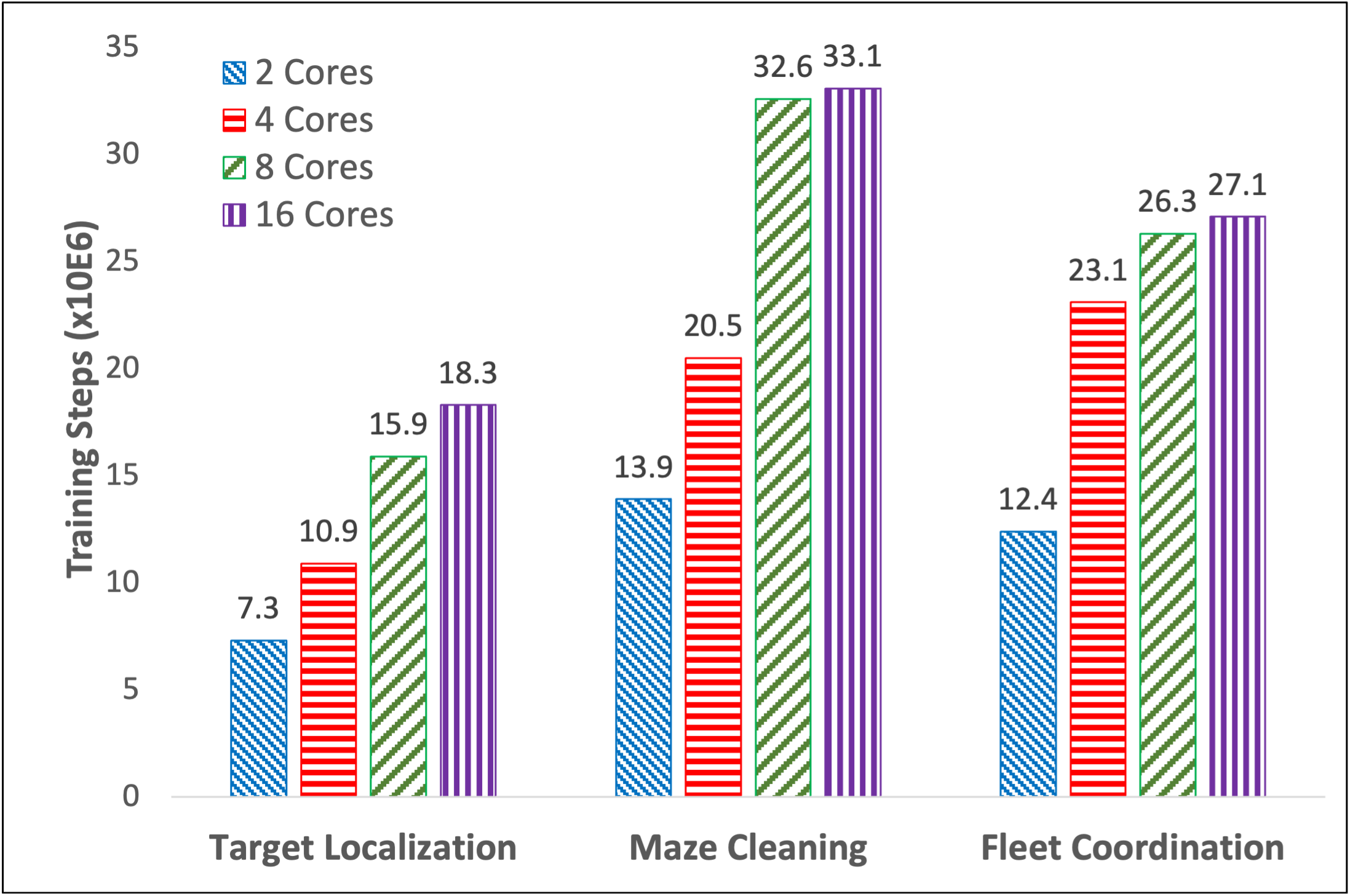}
         \caption{Training Steps vs CPU Cores}
         \label{Fig: Steps_CPU}
     \end{subfigure}
     \begin{subfigure}{0.486\columnwidth}
         \centering
         \includegraphics[width=\linewidth]{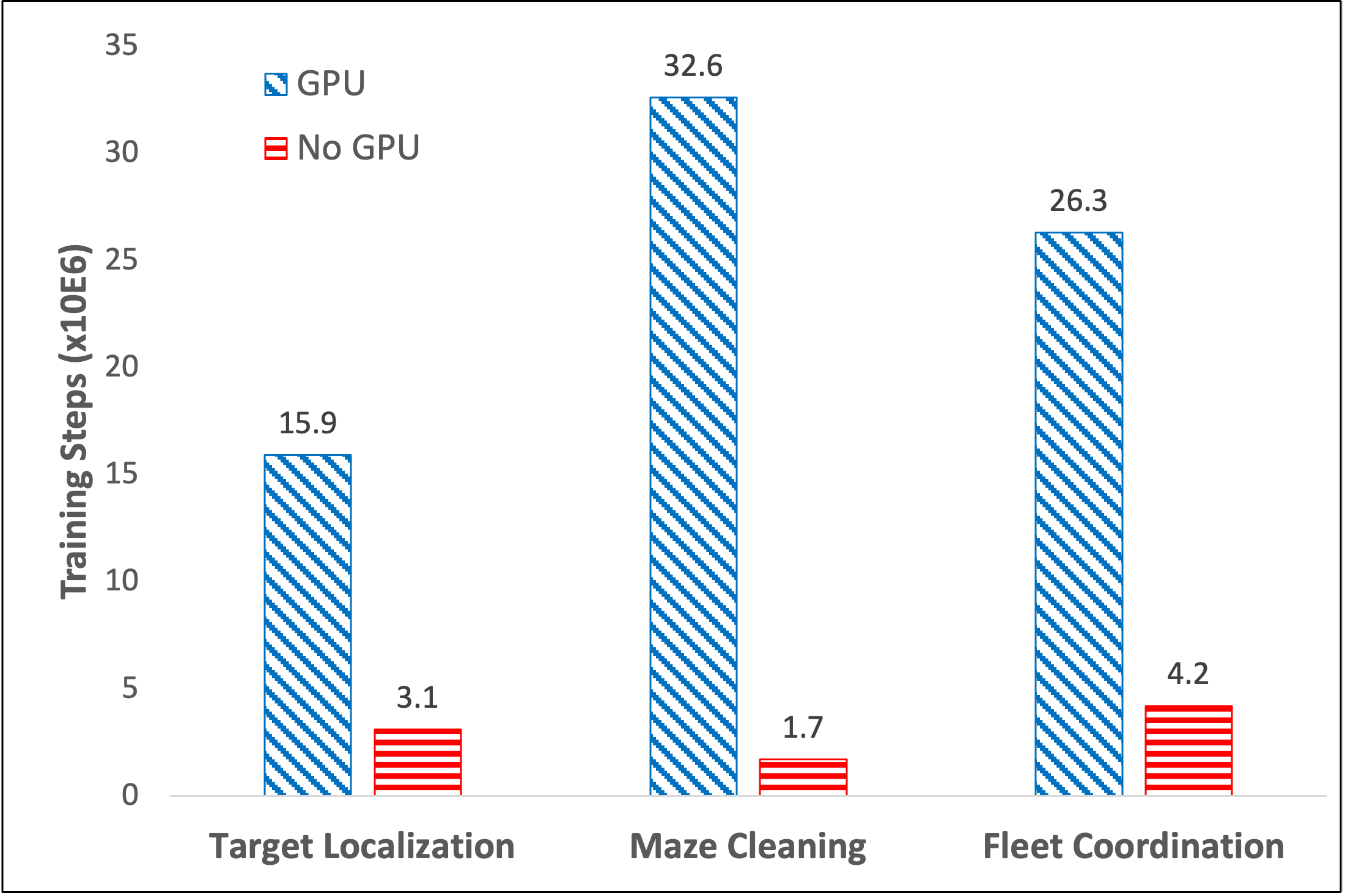}
         \caption{Training Steps vs GPU}
         \label{Fig: Steps_GPU}
     \end{subfigure}
     
        \caption{The total number of training steps in a 12h duration, while (a) varying the number of CPU cores (parallelized DRL) and (a) varying the use of GPU.}
        \label{Fig: Steps_CPU_GPU}
\end{figure}

While the training convergence across different number of cores is similar for most applications, and slightly better with more cores for some other applications, the case is different in terms of wall time. Fig. \ref{Fig: Steps_CPU} elaborates on the number of training steps obtained in each of the DRL applications within 12 hours of wall time, for varying number of CPU cores. It can be seen that, despite the similar training convergence in Target Localization and Fleet Convergence environments (previously shown in Fig. \ref{Fig: CPU_Results}), the training time is significantly different. Specifically, a training method using 16 cores can execute 2.5, 2.4, and 2.2 times the training steps with 2 cores, for the Target Localization, Maze Cleaning, and Fleet Coordination environments, respectively. This means that, in the case of Target Localization for instance, the algorithm with 16 CPU cores needed less than half the wall time to converge when compared to the algorithm with 2 CPU cores. This shows the importance of the CPU capabilities attribute used to assess candidate workers. It is also essential to point out that the effect of CPU cores on the wall time begins to saturate as the CPU cores increase, which can be shown when going from 8 to 16. This verifies the discussion in Section \ref{subsection DRL Training}, which states that at a certain point, further parallelization does not bring more benefit, and justifies the use of the $tan^{-1}$ function in the assessment process.

In terms of GPU, Fig. \ref{Fig: Steps_GPU} presents the effect of training with and without GPU, for the 3 different applications. GPUs are essential when training DNNs, especially when dealing with CNNs. As can be seen in the figure, training with GPU for a duration of 12 hours executes up to 19 times the training steps executed without GPU, which validates the consideration of GPU capabilities in the worker assessment process.

\subsubsection{DRL Model Sharing}

\begin{figure}[H]
     \centering
    \begin{subfigure}{0.485\columnwidth}
         \centering
         \includegraphics[width=\linewidth]{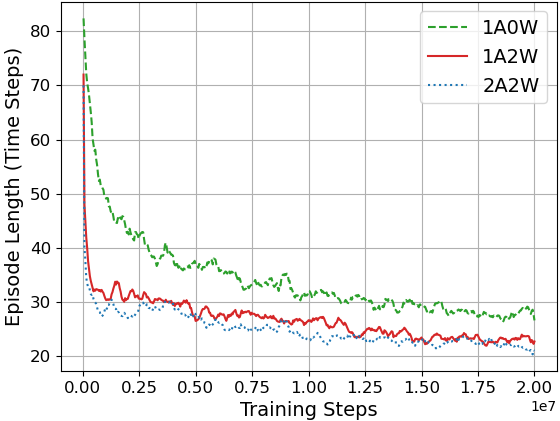}
         \caption{Target Localization (3A3W)}
         \vspace{1em}
         \label{Fig: TL_Sim}
     \end{subfigure}
     \begin{subfigure}{0.49\columnwidth}
         \centering
         \includegraphics[width=\linewidth]{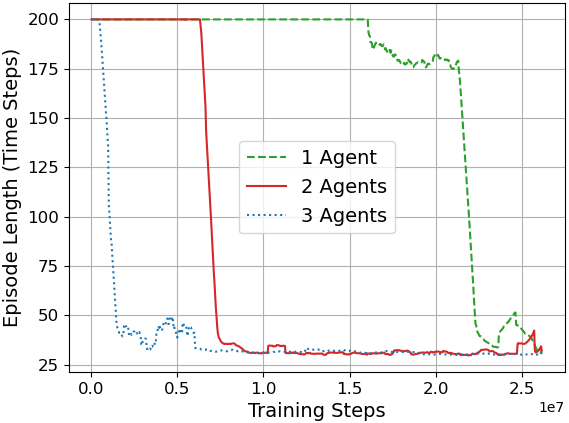}
         \caption{Maze Cleaning (5-agent)}
         \vspace{1em}
         \label{Fig: MC_Sim}
     \end{subfigure}
     \begin{subfigure}{0.49\columnwidth}
         \centering
         \includegraphics[width=\linewidth]{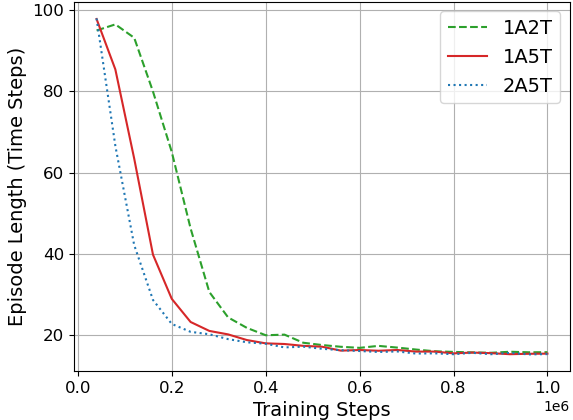}
         \caption{Fleet Coordination (3A10T)}
         \label{Fig: AV_Sim}
     \end{subfigure}
     
        \caption{The effect of model similarity on the learning performance, when training a 3-agent 3-wall target localization problem (3A3W), a 5-agent maze cleaning environment, and a 3-agent 10-target fleet coordination problem (3A10T).}
        \label{Fig: Sim_Results}
\end{figure}

To study the importance of the similarity metric proposed in Eq. \ref{eq: Similarity} for the DRL model sharing tasks, Fig. \ref{Fig: Sim_Results} shows the learning performance when using different expert models in assisting DRL, using Demonstration Cloning (DC) \cite{alagha2023multi}. In DC, expert models help current agents collect better experiences with more exposure to reward values, resulting in better learning. For the target localization problem (Fig. \ref{Fig: TL_Sim}), a team of agents is to be trained on environment with 3 agents and 3 walls (3A3W), with the help of 3 expert models that have been previously trained on different environments, including 1A0W, 1A2W, and 2A2W. It can be seen that the closer the expert model is to the current environment of interest (3A3W), the better the learning and the faster the convergence. Specifically, the expert model from the 2A2W environment assists the training the best, since it is the closest to the current environment. Similarly, for the maze cleaning problem, an expert model trained on a 3-agent environment provides the best assistant to train a 5-agent environment, when compared to expert models trained on single- and two-agent environments. For fleet coordination, an expert model trained on an environment of 2 agents and 5 targets (2A5T) gives the best assistant when training an environment of 3 agents and 10 targets (3A10T), when compared to expert models trained on 1A2T and 1A5T environments.

\subsubsection{Recruitment Optimization}
\begin{figure}[H]
    \centering
    \includegraphics[width=0.55\columnwidth]{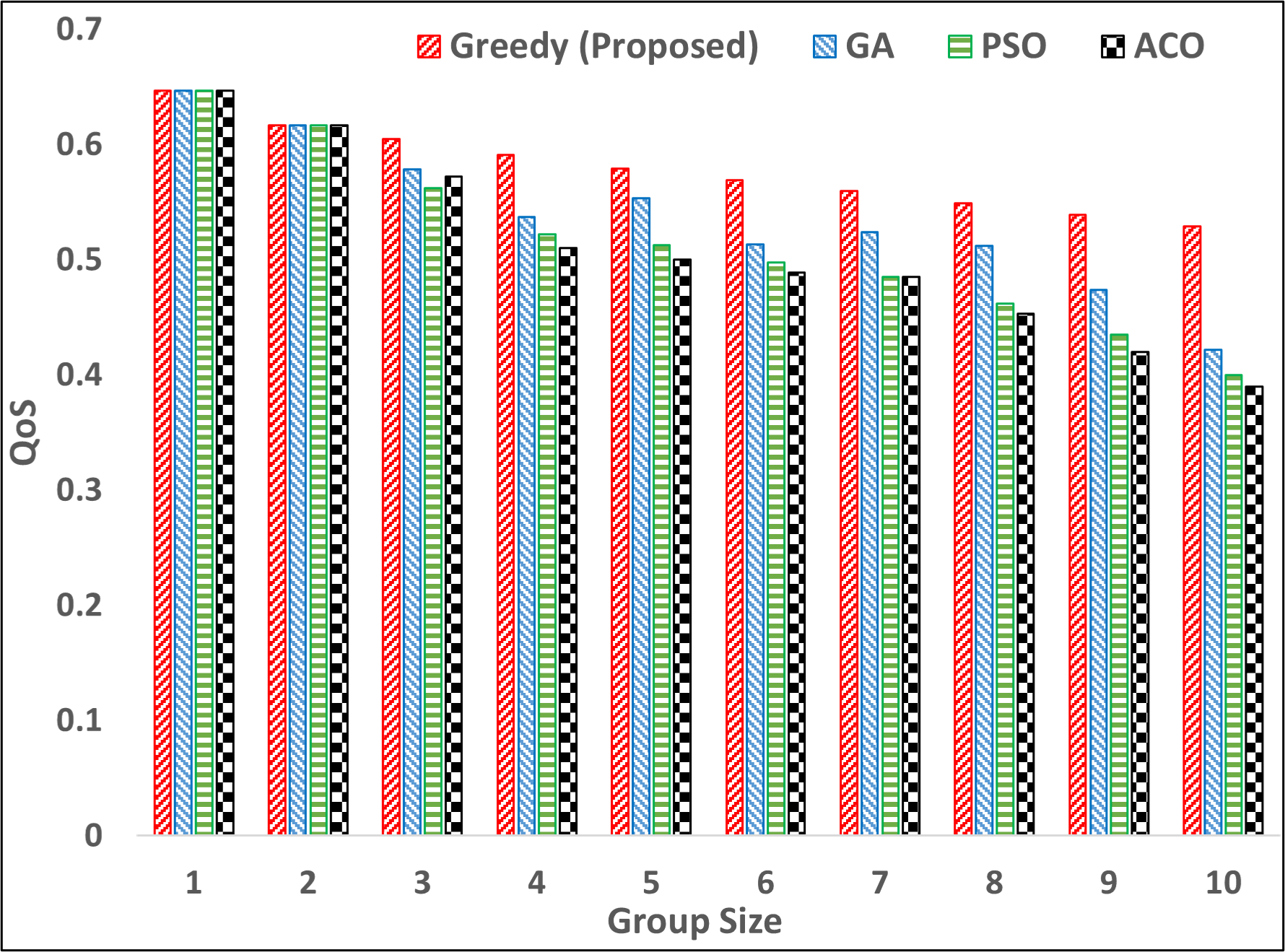}
    \caption{Comparison between the proposed greedy-based recruitment and the benchmarks for different group sizes.}
    \label{fig:QoS}
\end{figure}

To validate the choice of the greedy algorithm for the optimization of the recruitment process, the proposed method is benchmarked against common methods in crowdsourcing works, such as Genetic Algorithm (GA) \cite{alagha2019data}, Particle Swarm Optimization (PSO) \cite{wang2020worker}, and Ant Colony Optimization (ACO) \cite{wang2020method}. In GA-based methods, a population of candidate solutions (i.e. possible sets of workers) is created and iteratively modified through genetic operators such as crossover, mutation, and selection, aiming to converge toward an optimal solution eventually. In PSO-based methods, a swarm of candidate solutions (particles) explores potential worker selections based on their QoS values. Here, the positions and velocities of the particles are iteratively updated based on their own experience (personal best) and the experience of the swarm (global best). In ACO-based methods, the recruitment problem is modeled as a graph where nodes represent workers and edges represent possible selections. Ants placed on the graph deploy pheromones on the paths they take, which is proportional to the quality of the solution. Fig. \ref{fig:QoS} compares the performance of the greedy-based recruitment method with the benchmarks in terms of group average QoS for different group sizes, where a group size is the number of workers recruited. The results are obtained using a synthetic dataset of 600 candidate workers, where the workers' attributes are generated randomly following a uniform distribution for fair comparison. It can be seen in the figure that, regardless of the group size, a greedy-based method always outperforms the other benchmarks. This is mainly because the selection of workers occurs independently for each worker in a greedy method. Hence, the size of its search space is simply the available pool of workers, making the search process simple. On the other hand, GA-, PSO-, and ACO-based methods operate on a broader search space by considering the different combinations of groups of workers, making the search problem harder. It is worth mentioning that for the current results in Fig. \ref{fig:QoS}, on average, the greedy algorithm is nearly 200 times faster than GA, 73 times faster than PSO, and 132 times faster than ACO. This is significant for the deployment of the proposed framework on the blockchain.

\begin{figure}[H]
     \centering
    \begin{subfigure}{0.5\columnwidth}
         \centering
         \includegraphics[width=\linewidth]{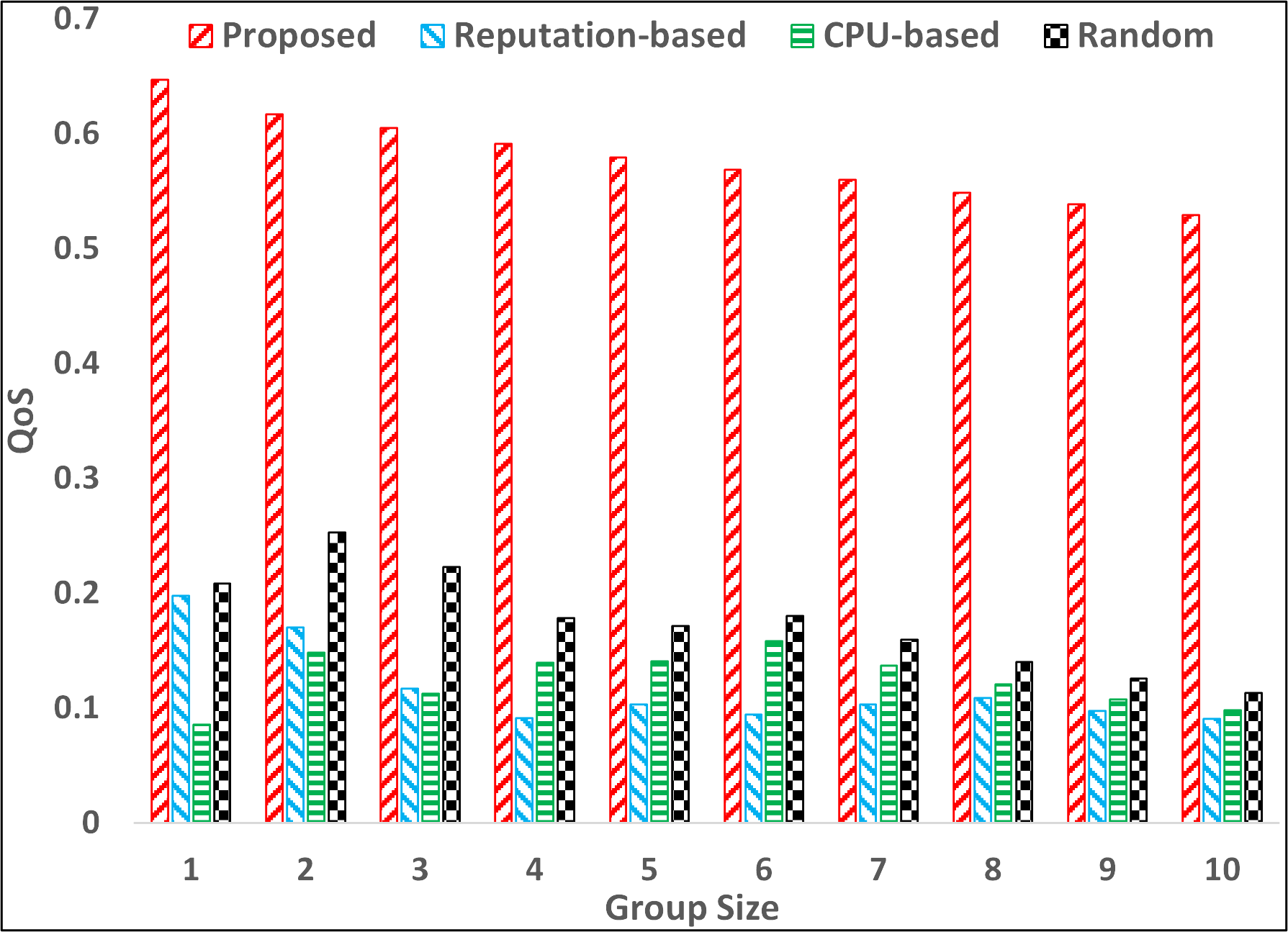}
         \caption{}
         \vspace{1em}
         \label{Fig: QoS-Bench}
     \end{subfigure}
     \vspace{-0.3em}
     \begin{subfigure}{0.55\columnwidth}
         \centering
         \includegraphics[width=\linewidth]{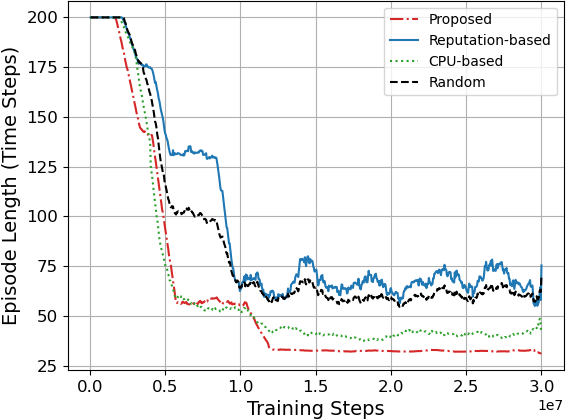}
         \caption{}
         \vspace{1em}
         \label{Fig: MC-Bench}
     \end{subfigure}
     \vspace{-1em}
        \caption{Comparison between the proposed method and different benchmarks in terms of (a) QoS for different group sizes and (b) DRL training results for a group size of 4, using the maze cleaning environment.}
        \label{Fig: QoS-RL-Benchmarks}
\end{figure}

To analyze the effect of the recruited workers on the QoS and the DRL training results, Fig. \ref{Fig: QoS-RL-Benchmarks} compares the proposed work based on the QoS in Eq. \ref{QoSTrainFormula} with several recruitment benchmarks. Since no works in the literature address the crowdsourcing of DRL training tasks, we choose benchmarks from similar domains in crowdsourcing and federated learning. Reputation-based recruitment methods \cite{abououf2021machine} focus on reputation-related attributes to assess the workers, with the aim of selecting the most reputed ones. CPU-based recruitment methods \cite{wehbi2023fedmint, chahoud2023demand} focus on the computational capabilities of the workers in terms of CPU and RAM. We also consider a random recruitment method as a baseline. As shown in Fig. \ref{Fig: QoS-Bench}, the three benchmarks fall short in terms of QoS when compared to the proposed work. This is mainly because each of the benchmarks considers only a subset of the DRL-related parameters considered in Eq. \ref{QoSTrainFormula}. To demonstrate the effect of the recruitment of the DRL task, Fig. \ref{Fig: MC-Bench} shows the training results (in terms of episode length to be minimized) using the workers selected by each benchmark. Here, the aforementioned 600-candidate workers dataset is used, and selected workers are given the task of training a model to solve a maze-cleaning environment of 4 agents. For a fair comparison, we modify our method to match each of the benchmarks by keeping only the attributes considered by them. As can be seen in the figure, reputation-based and random recruitment benchmarks struggle with convergence. This is due to the lack of DRL-related attributes such as computational capabilities. On the other hand, the CPU-based recruitment benchmark performs nearly as well as the proposed work, with the latter showing a slight performance advantage in terms of episode length. It is also worth noting that the workers chosen by the proposed work are nearly 20 times faster than the CPU-based benchmark in training the $30M$ steps, which is due to the use of GPU that is missing from the benchmark.

\subsubsection{Blockchain and Smart Contracts Complexity Analysis}
To analyze the complexity and feasibility of the proposed smart contracts, Table \ref{Table: gas} presents the gas cost of the deployment and execution of the smart contracts and their functions. Gas cost is defined as the computational effort required to execute operations such as smart contracts. In public blockchains, the gas cost can be used to determine the fees to be paid for executing transactions or running smart contracts based on the gas price. Since a Consortium Blockchain is proposed, the deployment and execution of the smart contracts do not require any payments by the users since the gas price is zero. However, the gas cost is a good measure of the complexity of the smart contracts and its functions to indicate their feasibility. As seen in the table, the gas costs are low, reflecting the feasibility of the proposed contracts. For reference, we present a benchmark of the gas cost of deploying and executing a similar UMC as discussed in \cite{kadadha2022chain}.

\renewcommand{\arraystretch}{1.2}
\begin{table}[h]
\caption{Blockchain gas cost.}
\vspace{-1.3em}
\begin{center}
\begin{tabular}{|P{0.35\columnwidth}|P{0.3\columnwidth}|P{0.15\columnwidth}|}
\hline
\rowcolor{Gray}
\textbf{Contract} & \textbf{Function} & \textbf{gas cost}\\ 
\hline
\multirow{4}{*}{UMC} & deployment & 735736 \\
\cline{2-3}
& \textit{addWorker()} & 85455 \\
\cline{2-3}
& \textit{addRequester()} & 74569 \\
\cline{2-3}
& \textit{updateStatus()} & 37286 \\
\cline{2-3}
& \textit{updateInfo)} & 67486 \\
\cline{2-3}
& \textit{getWorkers()} & 95287 \\
\hline
\multirow{4}{*}{TMC} & deployment & 1477451 \\
\cline{2-3}
& \textit{addTask()} & 487452 \\
\cline{2-3}
& \textit{allocateTask()} & 1053842 \\
\cline{2-3}
& \textit{updateTaskStatus()} & 99374 \\
\cline{2-3}
& \textit{submitOutcome()} & 29374 \\
\hline
\multirow{4}{*}{MMC} & deployment & 1357425 \\
\cline{2-3}
& \textit{addModel()} & 634341 \\
\cline{2-3}
& \textit{allocateModel()} & 903842 \\
\hline
\multirow{2}{*}{UMC - Benchmark \cite{kadadha2022chain}} & deployment & 1228566\\
\cline{2-3}
& \textit{addUser()} & 352352\\
\hline
\end{tabular}

\end{center}
\label{Table: gas}
\end{table}

It is worth mentioning that consortium blockchains are generally known to have a significantly lower latency and a higher throughput when compared to public blockchains. This is mainly because they have a predetermined set of nodes that are known to each other, which allows faster transaction processing due to optimized consensus mechanisms with a small number of trusted validators. The exact computations of latency and throughput depend on several factors such as the number of nodes, security requirements, and the consensus mechanisms, which are out of the scope of this work. However, since DRL task allocation in the proposed crowdsourcing system is not time-sensitive, studying the latency and throughput is not significant to this work. Nonetheless, in common consortium blockchains such as Quorum, the blockchain can handle hundreds to thousands of transactions per second (throughput) with latency as low as few milliseconds to 2 seconds.

\section{Conclusion and Future Directions}
\label{Conclusion}
In this paper, a novel blockchain-based framework for crowdsourced Deep Reinforcement Learning as a Service (DRLaaS) is proposed. The proposed framework aims at providing DRL-related services in terms of expertise and computational resources for users in the DRL domain. The proposed framework encapsulates two possible task types: DRL training and model sharing. In this framework, users submit tasks to be allocated to appropriate workers, who then return task outcomes as per the requirements. Specific recruitment metrics and processes are designed, aiming to find the best set of workers for a given task, while considering expertise, computational capabilities, reputation, and model similarity to the target environment. On a Consortium Blockchain, smart contracts are designed to manage worker recruitment and model sharing processes using Greedy methods. Experiments on several DRL applications, such as target localization, fleet coordination, and maze cleaning show the efficacy of the proposed recruitment metrics, especially in terms of computational capabilities and model similarity.

In addition to the contributions made in this paper regarding worker recruitment for crowdsourcing DRL training and model sharing tasks, there are several avenues for future research and improvements. One important direction involves incorporating quality control mechanisms to ensure the reliability and accuracy of the submitted outcomes. Another direction is to investigate diverse incentive mechanisms through game theoretic approaches to incentivize worker engagement and maintain participation. All the aforementioned is while considering the special nature of DRL tasks and their needs.

\bibliographystyle{model1-num-names}
\bibliography{bibliography}
\end{document}